\documentclass[sn-basic]{sn-jnl}%

\usepackage{graphicx}%
\usepackage{multirow}%
\usepackage{amsmath,amssymb,amsfonts}%
\usepackage{amsthm}%
\usepackage{mathrsfs}%
\usepackage[title]{appendix}%
\usepackage{xcolor}%
\usepackage{textcomp}%
\usepackage{manyfoot}%
\usepackage{booktabs}%
\usepackage{algorithm}%
\usepackage{algorithmicx}%
\usepackage{algpseudocode}%
\usepackage{listings}%
\usepackage{cleveref}
\usepackage{multirow}
\usepackage{url}
\usepackage{makecell}
\usepackage[normalem]{ulem}

\begin{document}

\title[Generative User-Experience Research for Developing Domain-specific Natural Language Processing Applications]{Generative User-Experience Research for Developing Domain-specific Natural Language Processing Applications}

\author*[1]{\fnm{Anastasia} \sur{Zhukova}}\email{anastasia.zhukova@uni-goettingen.de}

\author[2]{\fnm{Lukas} \spfx{von} \sur{Sperl}}\email{lukas.vonsperl@eschbach.com}
\equalcont{These authors contributed equally to this work.}

\author[2]{
\fnm{Christian E.} \sur{Matt}}\email{christian.matt@eschbach.com}
\equalcont{These authors contributed equally to this work.}

\author[1]{
\fnm{Bela} \sur{Gipp}}\email{gipp@uni-goettingen.de}

\affil[1]{\orgname{Georg-August University of Göttingen}, \country{Germany}}
\affil[2]{\orgname{Eschbach GmbH}, \country{Germany}}

\abstract{User experience (UX) is a part of human-computer interaction (HCI) research and focuses on increasing intuitiveness, transparency, simplicity, and trust for the system users. Most UX research for machine learning (ML) or natural language processing (NLP) focuses on a data-driven methodology. It engages domain users mainly for usability evaluation. Moreover, more typical UX methods tailor the systems towards user usability, unlike learning about the user needs first. This paper proposes a new methodology for integrating generative UX research into developing domain NLP applications. Generative UX research employs domain users at the initial stages of prototype development, i.e., ideation and concept evaluation, and the last stage for evaluating system usefulness and user utility. 
The methodology emerged from and is evaluated on a case study about the
full-cycle prototype development of a domain-specific semantic search for daily operations in the process industry. A key finding of our case study is that involving domain experts increases their interest and trust in the final NLP application. The combined UX+NLP research of the proposed method efficiently considers data- and user-driven opportunities and constraints, which can be crucial for developing NLP applications.}


\maketitle

\section{Introduction}
Natural Language Processing (NLP) has been recently extensively incorporated into industrial and domain applications. For example, NLP is used for speeding up processes, e.g., automation classification of types of customer feedback or filtering out spam emails, information extraction, e.g., named entity recognition to extract symptoms, diagnoses, and treatments from medical records, or auto-completing input forms with language models. Despite the broad integration, domain-specific NLP applications may require practicing more user-driven methodologies to address user needs with these applications.

Often, the data-driven approach falls short in exploring the needs of the domain users \citep{yang2018machine}. On the one hand, domain users are often integrated into development at the late test phase to evaluate the usability of ML/NLP applications \citep{dscoutUserResearch2019}. Unlike user-driven software development, the development of NLP applications depends mainly on data availability or experimenting with machine learning (ML)/NLP trends and thus is a major driver of application development. On the other hand, the user-driven development of a domain-specific ML/NLP application in medicine showed that close collaboration with the domain users in the earlier stages increases the effectiveness of the final product \citep{yang2017role}. Therefore, integrating user experience (UX) and human-computer interaction (HCI) research into ML/NLP research addresses users' needs, fuses their expertise, and increases intuitiveness, transparency, simplicity, and trust for the system users \citep{boukhelifa2018evaluation, paleyes2022challenges}. 

This paper reports on developing a real-world domain NLP application for improving effectiveness and efficiency in daily plant operations in the process industry. A major project goal was to achieve high user acceptance for the final application, which required involving users from the initial design phase. That resulted in the proposed 
methodology for integrating generative research into a development cycle of domain NLP applications by combining data- and user-driven concept creation, prototype development, and prototype evaluation. Widely applied in design and software development, generative UX research is a methodology that aims to learn about domain users and their stories \citep{hanington2007generative, anderson_2023}. To increase the credibility and utility of the final application, generative research engages the users in the early stage of development.  

The proposed methodology combines data- and user-driven approaches to full-cycle UX-driven prototype development for NLP applications. Prototyping or developing the first minimal viable products (MVPs) most benefits from the methodology. The proposed methodology consists of the following phases. First, the methodology describes generative UX research steps to explore opportunities for ideas and application concepts. This phase relies on exploratory data analysis and contextual inquiries, i.e., user interviews and on-site observations, as a source of idea generation. Furthermore, the domain users validate a proposed concept and help define the most important features for prototype implementation. Second, the prototyping addresses the most important identified research challenges crucial for usability assessment. Lastly, the prototype evaluation measures user utility via contextual inquiries before and after interacting with the prototype. 

Our team comprises UX researchers, data scientists, software developers, and a project leader because such a team composition facilitates a deeper understanding of the capabilities and limitations of the domain NLP application \citep{dove2017ux}. Therefore, the paper may have specific interest for the individual practitioners with the above-mentioned roles or similar teams. Moreover, practitioners of user-driven research, e.g., customer research departments and quality assurance officers, could benefit from the proposed methodology and reported results. 

The paper is structured as follows. \Cref{sec:rel_work} briefly reviews previous efforts and UX methodologies in creating ML applications. \Cref{sec:methodology} describes a methodology of generative UX research, in which we propose to optimize user experience in the usually data-driven development of domain NLP applications. Our case study's methodology covers a full-cycle NLP prototype development from generative research for concept ideation to user-driven prototype evaluation. \Cref{sec:case_study} reports a case study about implementing the proposed methodology to prototype an industrial NLP application for the process industry. \Cref{sec:discussion} discusses the impact of the generative UX research on developing our NLP prototype and summarizes the findings into a reusable methodology for applying generative research to increase the quality of NLP domain applications.

\section{Related Work}
\label{sec:rel_work}

Domain-driven design (DDD) is a software development methodology that understands, conceptualizes, and models complex domains into applications closely aligned with business needs \citep{evans2004ddd}. The core of DDD lies in close interaction between developers,  stakeholders, and users to study and implement the business logic of the underlying processes. According to DDD, the building blocks of software should reflect the physical world as closely as possible in terms of both functionality and naming (e.g., \citep{millett2015patterns, khononov2021learning, Vernon13, hippchen2017, hippchen2019}). The methodology emphasizes understanding a domain by understanding and utilizing its ubiquitous language by all development participants in all planning, development, and refinement stages. Moreover, DDD enables dynamic adjustments of a domain model to changes in real life through an iterative process of collecting feedback from domain users.

User experience research has become a requirement for practical and effective software development in addition to the domain understanding \citep{Fronemann2014, law2010modelling}. 
UX puts perspective onto users, thus defining functionality and look of new products or features and ensuring a need in new products \citep{kuniavsky2003observing}. UX research methods analyze how users interact with a system of interest and explore what users miss or find inconvenient. Quantitative (e.g., questionnaires) and qualitative (e.g., interviews) methods are used to collect and analyze user feedback \citep{rohrer2014use}. Structured interviews with domain-specific probe questions help UX elicit knowledge about the domain essential for better understanding system requirements \citep{hoffman1995eliciting}.

The rapid growth of ML applications evolved from automating routine processes (e.g., spam filtering) to user-centered applications (i.e., automatic information extraction based on a specific user request). Although ML applications put users into the loop to ensure the usability of the application \citep{yang2017role, dove2017ux}, most of the ML user-oriented development is product-focused, i.e., the rather later stages engage users to evaluate product usability \citep{visser2005contextmapping}. UX-driven research embeds user perspective evenly during the development of ML-powered applications, i.e., at the stages of requirement understanding \citep{carmona2018relationship}, data collection \citep{Park2021}, implementation \citep{Carney2020}, and evaluation \citep{cambo2018user}. The UX research explores multiple user interaction and engagement aspects in the development process. UX research increases the effectiveness and communicability of ML applications according to the user's mental maps \citep{agner2020recommendation}, creates workflows to enable effective joint creativity process of users and ML models \citep{kayacik2019identifying}, ensures the best user experience from the services \citep{dove2017ux}, incorporates expert knowledge into improving the ML applications \citep{amershi2014power,dudley2018review}, and engages users into accuracy evaluation \citep{cambo2018user,boukhelifa2018evaluation}.

Although users participate in multiple steps of the development process, a deeper user-centered perspective is required to distill users' real needs \citep{dove2017ux}. Most UX studies for ML applications follow a one- or two-step product-centered process \citep{visser2005contextmapping, yang2018machine}: (1) interact with the users to learn about a domain, (2) perform usability evaluation of an ML application. Very little research integrates the domain users into the first step to explore their perspectives for idea generation and concept analysis. Exploring user perspectives would include learning about the domain users, their roles in this domain, how they use a currently existing system (if any), and what needs and pain points the users may have. Analysis of this information could evolve the ML development into a user-centered viewpoint.

Generative research, often applied in design and software development, aims to learn domain user stories and personal experiences through exploratory, generative, and evaluative methods \citep{hanington2007generative}. Generative research targets a deeper understanding of user needs and develops product concepts based on the discovered user goals, utility, and underlying reasons for their actions \citep{anderson_2023}. 
Generative research is conceptually close to the practical design thinking methodology often applied in industry \cite{brown2008design, Lindberg2011, dellera2020}. Although the ideation process in both methodologies is user-centered, generative research focuses more on exploring domain users in their environment and deriving prototype ideas based on immersive domain learning.

\section{Methodology}
\label{sec:methodology}

In this section, we propose a methodology of generative UX research for optimizing a data-driven domain-specific NLP application toward higher user satisfaction.
Generative research encourages user-focused development by getting to know the users better and explores opportunities for an ML/NLP application by learning users' stories, pain points, and needs \citep{visser2005contextmapping, anderson_2023}. 
The methodology is grounded on a domain-driven design (DDD) \citep{evans2004ddd} of deep domain understanding through close collaboration between developers and domain experts. The iterative knowledge exchange about domain know-how, problem-solving requirements, and implementation constraints leads to effective domain modeling, i.e., a conceptual representation of the domain at consideration. An effective and knowledge-rich domain model reflects internal and external business processes most fully. Moreover, a model should be clear to both domain experts to provide feedback and developers to act as a foundation for implementation. The generative research puts domain experts even more into perspective by switching from business processes to domain experts' workflows and their daily routines.

Although generative research has already been employed in design and software development, it lacks coverage of full-cycle development of ML/NLP applications. For example, \cite{anderson_2023} focused on how to learn about user needs and experiences. \cite{hanington2007generative} explored how to create a product concept based on the learned information from users and employed the users to evaluate this concept. Similarly, design thinking proposes a user-centered process of iterative ideating and prototyping \citep{brown2008design, Lindberg2011, dellera2020}. \cite{dscoutUserResearch2019} employed users to evaluate a prototype before developing a product for the production-ready state.
Moreover, these methodologies do not consider an aspect of data-driven research that should be considered when creating an NLP application. 
The proposed methodology combines these previous efforts on generative research, design thinking, and domain understanding through exploratory data analysis required in ML/NLP applications.

\begin{figure}[ht!]
\centering
\includegraphics[width=1.0\textwidth]{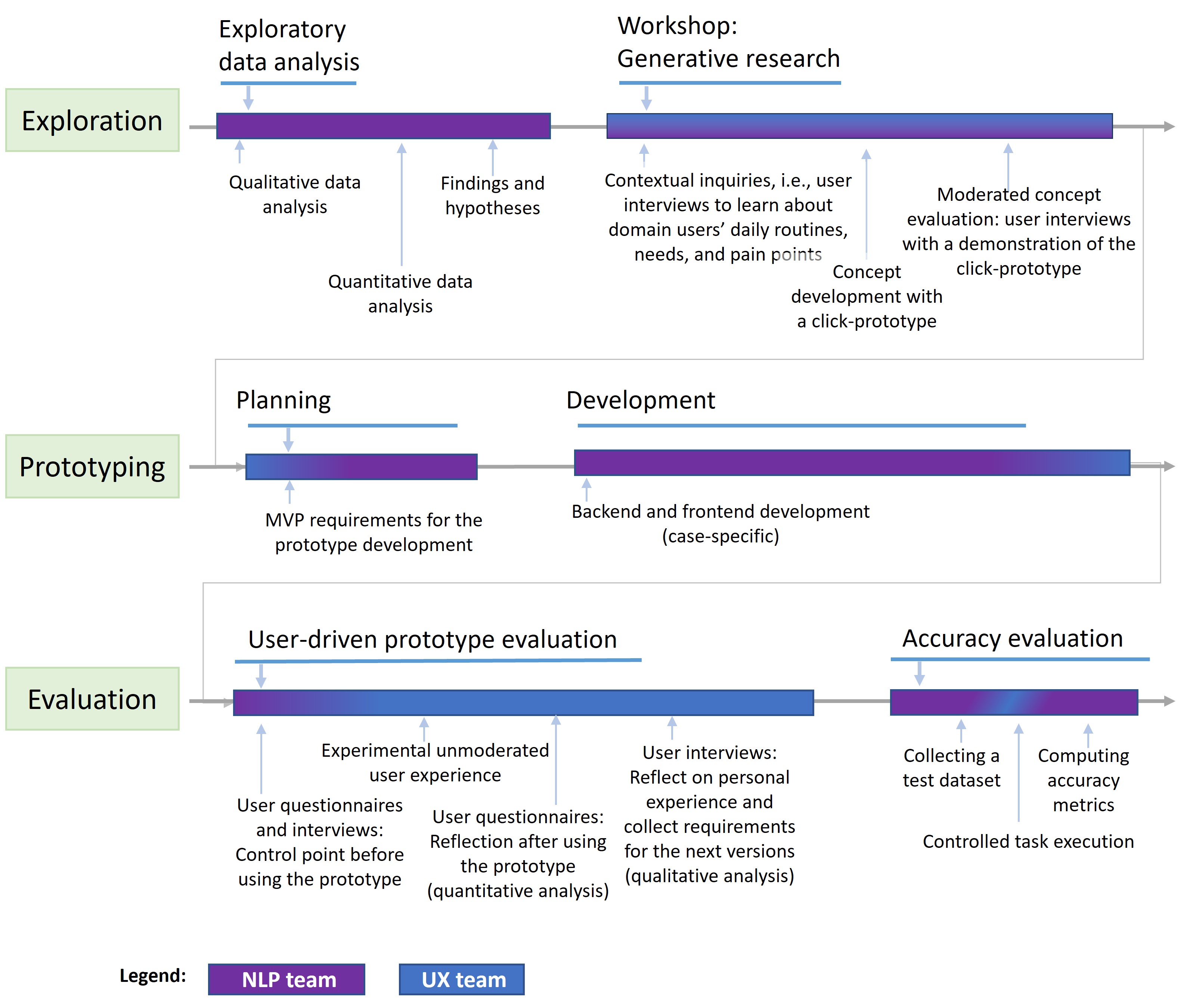}
\caption{A timeline for the proposed methodology of generative UX research for prototyping domain NLP applications used in our case study. The methodology consists of three stages: exploration (data exploration and generative research), prototyping (development of the prototype and evaluation (assessment of accuracy and user utility). 
The phases show the approximate share of the workload of the UX and NLP teams. For the majority of the phases, both teams need to be involved. Especially during the generative research phase, an equal involvement of the two teams enables deep domain understanding. } 
\label{fig:timeline}
\end{figure}

The methodology for full-cycle prototype development through generative research combines UX+ML/NLP to (1) explore ideas for an NLP application based on the stories of domain users, (2) collect feedback on a concept of a proposed NLP solution to their identified needs or pain points, (3) conduct a user-driven evaluation of a prototype from perspectives of user experience with a new NLP solution and accuracy estimation. \Cref{fig:timeline} depicts our proposed methodology to develop and evaluate a domain-specific NLP application. The methodology consists of three steps: 
\begin{enumerate}
    \item \textit{Exploration}, i.e., design ideas for a prototype through exploratory data analysis, interviews, and observation of the domain users in their environment.
    \item \textit{Prototyping} to develop a fully working prototype with back- and front-end.
    \item Prototype \textit{evaluation}, i.e., evaluation of the change in user experience and prototype accuracy assessment.
\end{enumerate}

Exploring ideas for an NLP prototype through \textit{generative UX research} aims to collect the necessary information to understand a domain in general, get familiar with the domain data, and learn about the user's needs. We tailor a methodology for concept creation via generative research proposed by \cite{hanington2007generative} to a more specific use-case of concept ideation for a domain-specific industry application. A proposed generative research for NLP consists of three steps: 
\begin{enumerate}
    \item \textit{Exploratory data analysis}: Learn about the domain system and data and create hypotheses we could probe during user interviews.
    \item \textit{Contextual inquiries}: (1) Conduct user interviews in a focus group to learn about their daily routines and systematize user interviews to identify patterns, and (2) Generate ideas to propose a concept for an NLP application or feature.
    \item \textit{Concept evaluation}: Perform moderated concept evaluation of a click-dummy, i.e., enable supervised interaction with a ``click-dummy'' prototype to test the usability and usefulness of the proposed concept.
\end{enumerate}

The user-driven prototype evaluation aims to assess prototype accuracy and how user utility changes after interacting with the prototype. Unlike concept evaluation at the stage of generative UX research, the prototype is evaluated in an open environment. The combination of the two perspectives (i.e., accuracy and user utility) determines the success level of the prototype in the first step toward a system for shared domain knowledge. The evaluation consists of three steps \citep{dscoutUserResearch2019}: 
\begin{enumerate}
    \item \textit{Control}: Use questionnaires and user interviews to identify an initial state of user experience, i.e., how users search for the information before independently interacting with the prototype;
    \item \textit{Experimental}: Provide the prototype to interact with it through provided tasks and openly throughout daily tasks as they feel fit;  
    \item \textit{Reflection}: Reuse the same questionnaire and interview scripts to identify how the user experience changes and learn if the system works as expected or surprises them and provides insights.
\end{enumerate}

The methodology combines data- and user-driven approaches that correspondingly originate from ML/NLP and UX research. A team of UX designers, data scientists, and developers ensures diverse and effective collaboration for each stage in the proposed pipeline \citep{yang2018investigating}. Discussing prototype ideas requires a balanced view of project aspects such as technical capabilities of NLP approaches and models, possible limitations (e.g., data or computation power limitations), and benefit to domain users \citep{yang2018machine, carmona2018relationship}. The ultimate goal of the methodology is to increase user involvement and interest in an industrial application and increase system usefulness \citep{paleyes2022challenges, boukhelifa2018evaluation}. 

\section{Case study}
\label{sec:case_study}
This case study describes implementing a domain-specific NLP application into a digital tool for capturing shift logs and managing plant operations in the chemical process industry. Shift logs usually contain problems and solutions arising during production, performed tasks, instructions from supervisors, reports about production goals, changes on production lines, and much more information about daily operations. As the system installed for a study participant plant had been used for several years, a large collection of text logs was available as a data source for NLP applications.

We consulted a user subgroup that closely interacts with the system daily to ideate how an NLP feature or application would look and evaluate which solution would be most effective for the domain system users. The main challenge was that the way the users intended to use the NLP application within the overall scope of tasks did not coincide with the data shape or form. Therefore, the goal of the case study was to devise a pilot idea based on the user perspective and derive an application that will assist domain users' daily tasks based on the data at hand. For this, we employed and implemented our proposed methodology of generative UX research for full-cycle prototyping domain NLP applications.

\subsection{Exploration}
\label{sec:exploration}

\subsubsection{Exploratory data analysis}
\label{sec:data_analysis}
A goal of domain understanding and exploratory data analysis is to learn about the area for the prototype and investigate data caveats that can either support or limit various stages of development. Our domain exploration steps included qualitative analysis, i.e., examining text data and workflows of domain experts, and quantitative analysis, i.e., statistical analysis of the text records and composition of domain vocabulary.

\paragraph{Qualitative analysis} 
We implemented an NLP feature for an existing system of electronic shift books, which is used for logging information about daily operations in the process industry, e.g., chemical or pharmaceutical production. Integrating a feature into an existing system involves conceptualizing difficulties, such as adhering to certain workflows and users' mind maps \citep{yang2018machine}. We conducted a case study with one plant in the chemical domain that operates in the German language. A database with the plant's daily operations logs is used for data analysis and prototype development.

\begin{figure}[ht!]
\centering
\includegraphics[width=\textwidth]{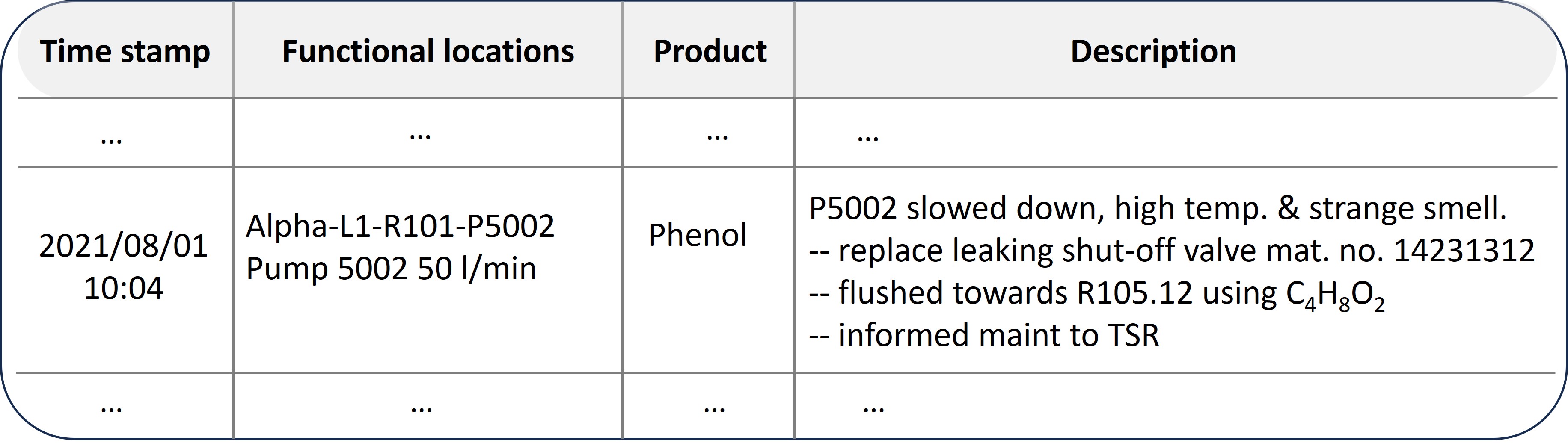}
\caption{A simplified example of a record in a logging system for the daily operations in the process industry. A record is semi-structured, i.e., it has structured attributes of the records and unstructured text descriptions. The domain language contains professional terms, abbreviations, shortenings, chemical formulas, various IDs (e.g., order IDs), and plant machinery and equipment inventory codes. The example is in English for better readability.} 
\label{fig:log}
\end{figure}

Records in shift books differ from regular text documents by their semi-structured nature, i.e., one record in a logging system contains both unstructured text fields and structured fields, which act as record attributes. \Cref{fig:log} depicts a simplified mock-up of the log structure. The assigned attributes can either complement the text data, e.g., text can report about an assigned functional location, or duplicate information in the text, e.g., a functional location is both mentioned in the text and is attached as an attribute. In some cases, the attributes remain empty, although they are mentioned in the text.

\begin{figure}[ht!]
\centering
\includegraphics[width=1.0\textwidth]{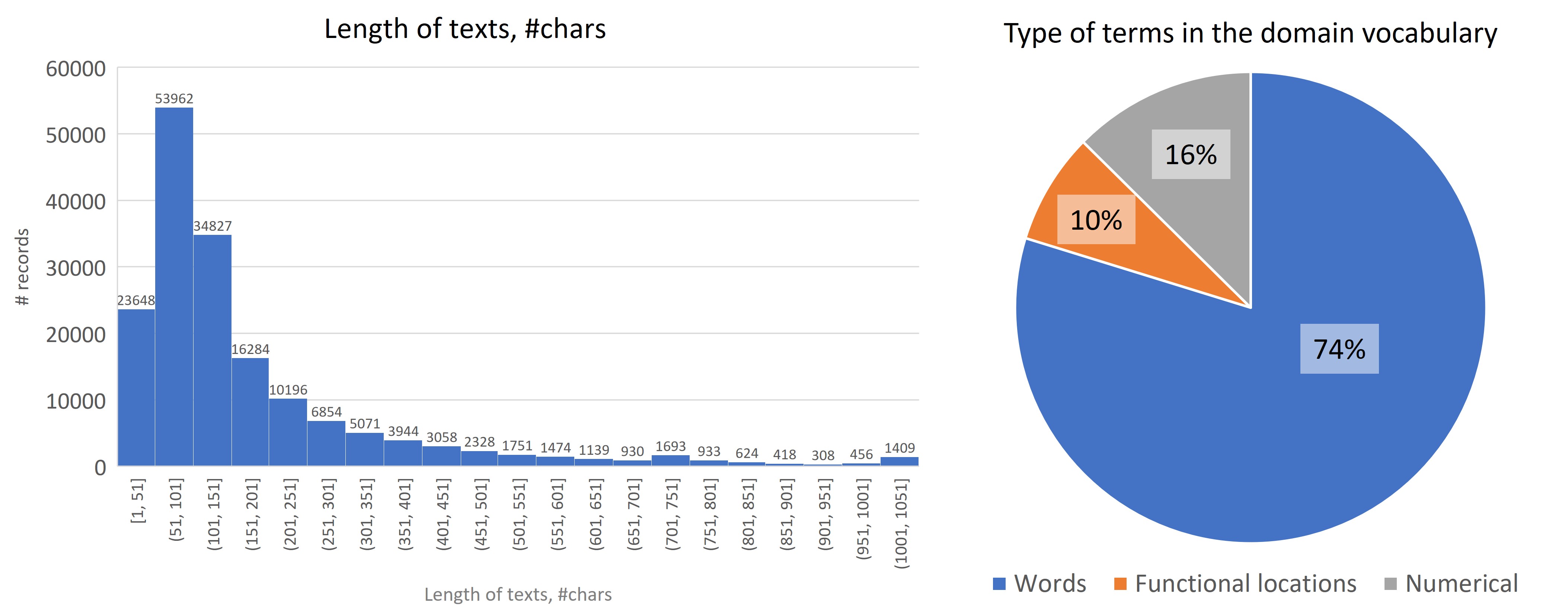}
\caption{Quantitative analysis of the domain text records of a logging system. Left: A histogram of the length of text records shows that most records are very short, i.e., smaller than 200 chars. Right: The domain vocabulary includes words and many codes of functional locations (i.e., machinery) that act like nouns. Additionally, the vocabulary contains a lot of digit-containing terms that refer to IDs or numerical values, e.g., measurements.} 
\label{fig:data_exploration}
\end{figure}

We observed two significant signs of a domain language. First, records contain quite a diverse writing style. While some records contained complete sentences, some looked like handwritten notes about statuses and a list of tasks to address later. Second, the text contains numeric data, abbreviations, product names, special symbols, terminology related to chemistry, and encoded terms that act like words. For example, \Cref{fig:log} shows that the terms ``P5002'' and ``R105.12'' act as nouns that correspondingly refer to a pump and a reactor. But outside of this domain, knowing the meaning of these terms is impossible. Moreover, the abbreviation TSR, which stands for technical safety review, and ``maint,'' which stands for ``maintenance,'' illustrate the short word forms used in this domain.

\paragraph{Quantitative analysis}
We performed quantitative data analysis to explore the qualitative analysis findings further and compare a general domain with the production-specific domain terminology. The analysis included the following tasks: (a) evaluate the length of the text records, (b) inspect text composition, and (c) explore the difference between the domain language and the general domain.

\begin{figure}[ht!]
\centering
\includegraphics[width=1.0\textwidth]{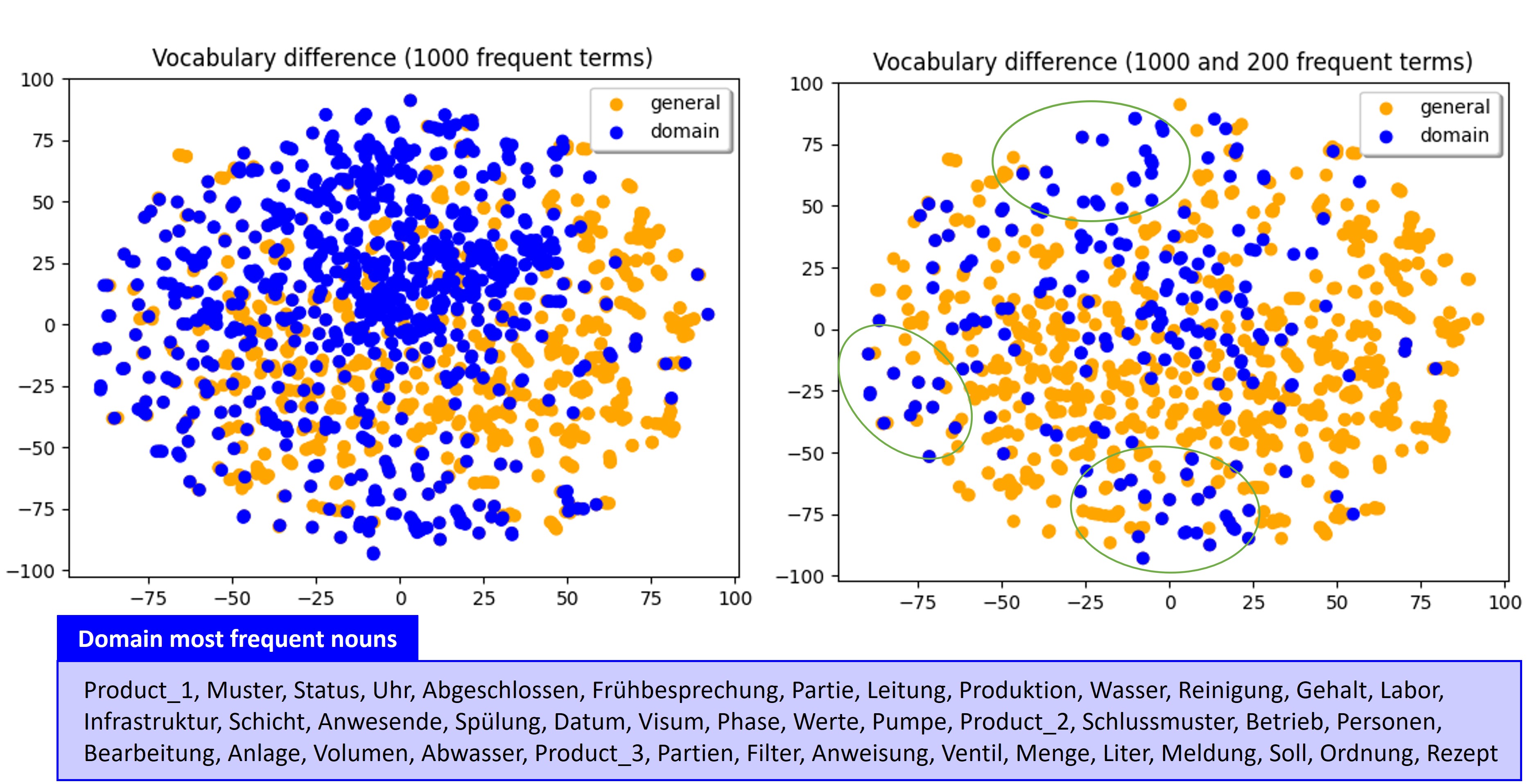}
\caption{Analysis of the domain language shows a significant difference to a general domain (Wikipedia): t-SNE projection of the vectorized domain and general vocabulary clearly shows spots with no overlap between the most frequent terms. Left plot: 1000 general and domain most frequent terms. Right plot: 1000 general and 200 domain most frequent terms. In the list of frequent domain terms, ``Product\_\#'' is an anonymized version of the real product names.  } 
\label{fig:vocabulary}
\end{figure}
 
\Cref{fig:data_exploration} shows a histogram of the text lengths, where most records do not exceed 200 chars, i.e., the size of Twitter posts. A pie chart depicts the proportion of the types of terms based on the overall number of terms in a dataset for analysis. We observed a significant impact of the numeric values (10\%) and encoded terms of the local domain (16\%), i.e., abbreviations for the functional locations used in the production. Such terms require local dictionaries to enable decoding the terms for out-of-domain understanding.

To explore the prevalence of the domain terminology, we visualized terms from the local domain and general use in a scatter plot. For the local domain, we extracted 1000 most frequent nouns from 20,000 randomly selected text records. For the general domain, we used a dump of German Wikipedia  \footnote{\url{https://github.com/GermanT5/wikipedia2corpus}} and extracted the 1000 most frequent nouns from the randomly selected 20,000 sentences. We focused on nouns as the most content-bearing terms. \Cref{fig:vocabulary} depicts a t-SNE projection of all extracted terms vectorized with fastText word embeddings \citep{mikolov-etal-2018-advances} and shows that domain and general terms lean towards different parts of the plot. Moreover, on the right plot, we see that some frequent terms are unique and do not overlap with the general domain. For example, the terms ``Trennschichtsonde,'' ``Einstiegsbereit,'' and ``Laderohr'' do not occur in Wikipedia.

\paragraph{Findings and hypotheses}
Our qualitative and quantitative analyses led to the following findings: (1) Text quality was quite uneven compared to the regular text, i.e., short texts of Twitter size often have the format of handwritten notes with incomplete sentences. (2) The text data contained many domain terms, i.e., specific terminology for the process and chemical industry, domain abbreviations, product names, and codes of functional locations that act like nouns. We hypothesized that text records with such uneven structure and writing style most likely have a short lifetime and are mainly used to keep track of the current processes and production state rather than as a source of knowledge collected over the long term.

\subsubsection{Workshop: Generative research}
\label{sec:exploration_interviews}
The objective of generative research is to learn how domain users use the system daily and elicit ideas for an NLP-based solution to improve the effectiveness of their existing processes. We performed contextual inquiries, i.e., one-to-one interviews with representative domain users in their environment \citep{beyer1999contextual}. We organized a week-long workshop at a plant to ensure an immersed environment for the research team \citep{soini2005workshops, beyer1999contextual}. 
The workshop's goal was to better understand experts' processes and workflows through observation and interaction with them \citep{evans2004ddd}. The observed processes were to be organized into explanatory models to be later distilled into ideas for the prototype implementation.

\paragraph{Contextual inquiries: User interviews}
The guided interviews consisted of open-ended questions to learn about the stories of the interviewees' daily routines. Interviewees in a focus group have expert knowledge of the logging system and represent most of the roles interacting with it, thus ensuring different perspectives of the contextual inquiries. The focus group consisted of seven persons with multiple roles in production: a plant coordinator, three shift leaders, a laboratory manager, a production assistant, and a second plant manager. 

The interviews followed two objectives: first, to learn how the interviewees use the logging system during their shifts, and second, to observe if they challenge our hypotheses derived after the exploratory data analysis. The personal interviews contained the following questions and inquiries: 
\begin{enumerate}
    \item Please tell us about your shift. Which events do you usually have during your average work day?
    \item How do you use the logging system in your daily tasks? On which occasions? What information would you be interested in there?
    \item How would you describe the quality of the logged text? Is there any information that you may lack?
\end{enumerate}

We summarized the interviews as follows: (1) a timeline of a shift that generalizes three stories of the shift leaders, (2) an overview of answers to the last two questions from all interviewees. These forms of summarization followed two different goals. First, we need to find the patterns in the user stories and identify which challenges or problems the interviewees mention at each step that we can turn into ideas for an NLP application. Second, to get an overview of the diverse opinions that can enable a broader picture for generating ideas.

\begin{figure}[ht!]
\centering
\includegraphics[width=1.0\textwidth]{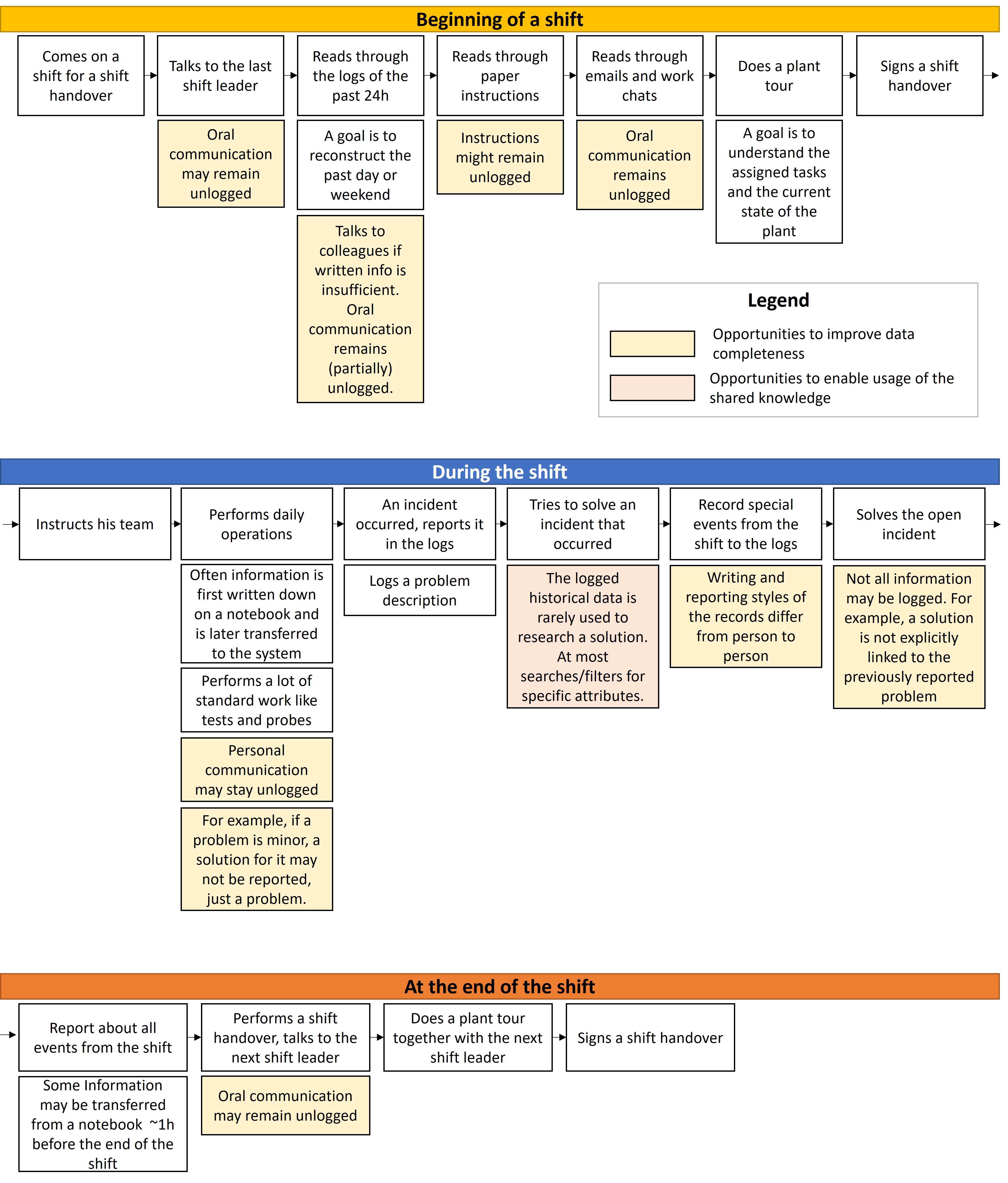}
\caption{A generalization of a shift from a shift leader resulting from our interviews that acts as a domain model for our user study. The context inquiries identified three phases of a shift and the main tasks and events that may occur during the day. We provide a short description and outline the challenges of the steps in all phases, which we can turn into opportunities for an NLP application.} 
\label{fig:shift}
\end{figure}

\begin{figure}[ht!]
\centering
\includegraphics[angle=90,height=17cm]{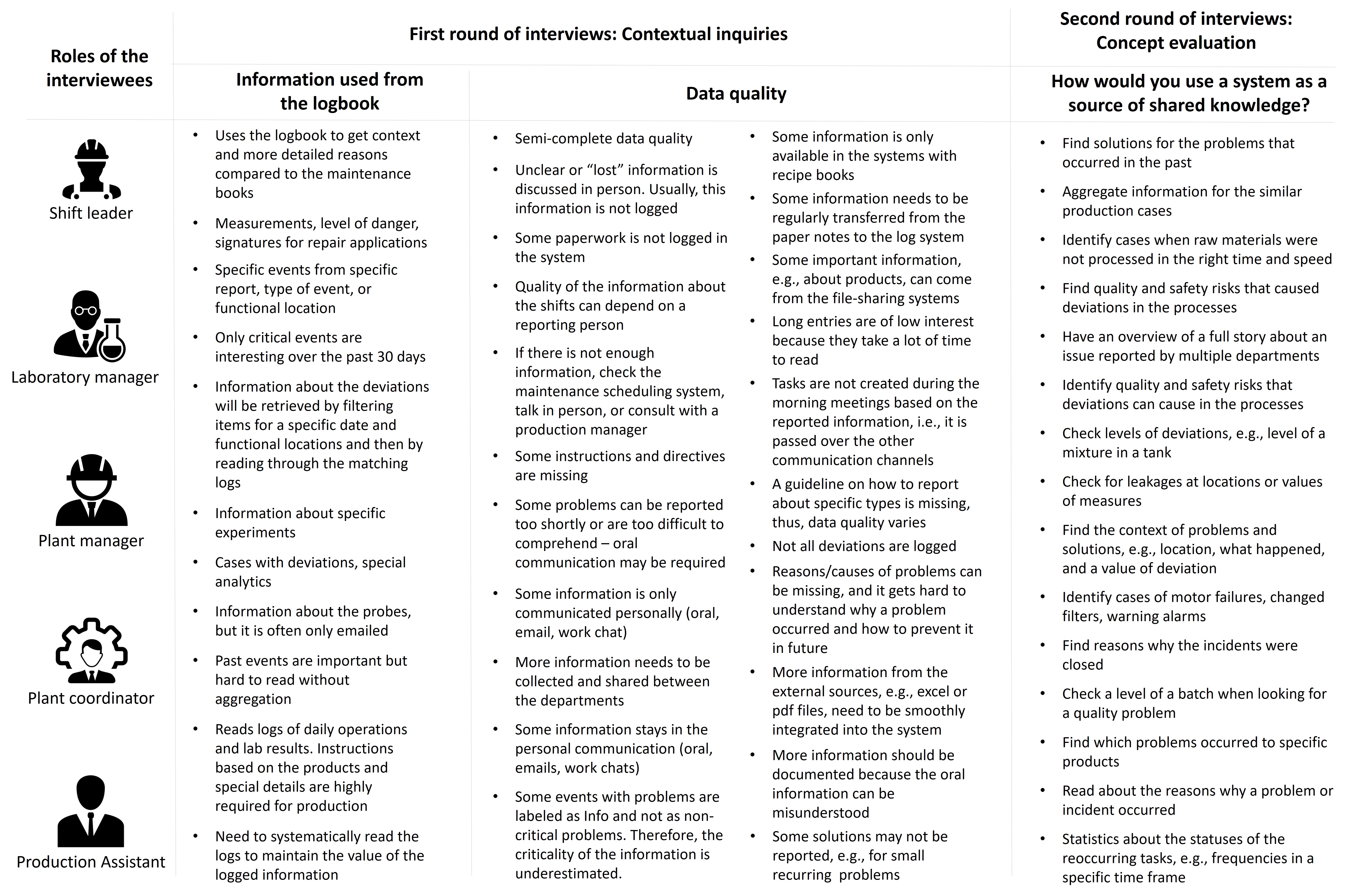}
\caption{An overview of the user replies for the questions from two rounds of interviews: to help create an idea for an NLP prototype and evaluate a proposed prototype concept. From the interviewees with different roles, we learned about the diversity of use cases of the logging system. Although the interviewees described a few cases of data incompleteness, they agreed that using the information from the logged records as shared knowledge can support them daily.} 
\label{fig:interviews}
\end{figure}

\Cref{fig:shift} generalizes an average shift of a shift leader. We identify three phases of a shift, i.e., beginning, middle, and end of the shift, which determine which tasks a shift leader needs to perform. For the steps in the phases, we indicated a motivation for a step and which mentioned aspects we could turn into opportunities. For example, at the beginning of a shift, a shift leader reads through the logs of the past 24 hours to collect an overview of what happened when they were not on duty. We learned that if not enough information was logged, they talked to their colleagues, and the information from the oral communication remained unrecorded. We noted that recording the additional information from communication was an opportunity to improve data completeness in the logging system. We identified multiple cases throughout a shift, which indicated that more data could be logged to assist in more informed daily operations. Additionally, we learned that shift leaders rarely use the system to research a solution for solving incidents. The most common way to find information of interest is to read through the recent records or use a string-match search to find specific keywords, e.g., product names or order IDs. This aspect would be an opportunity to facilitate knowledge sharing in the logging system.

\Cref{fig:interviews} overviews how the interviewees use the logging system and their feedback about the data quality. On the one hand, the interviewees indicated a diverse spectrum of information they use from the system, e.g., acquiring more context to the information from other systems running in the production, staying informed about deviation cases, reading about results of specific experiments, or learning only about critical events over the past 30 days. On the other hand, the interviewees agreed on the problems in the data quality and described it as semi-complete. For example, the interviewees revealed that they often require oral communication to clarify unclear logged information. Additionally, they may have needed to refer to external sources of information to understand the full picture (e.g., emails, work chats, or attached files). Moreover, the reporting style was person-dependent and sometimes lacked a reporting standard for specific events. The diverse writing styles may have depended on the type of reported event, i.e., reporting about some events required a higher level of detail, whereas reporting about others required only short messages due to using other systems for these events.

The interviews confirmed our hypotheses about the high data quality variation and the logged information's short lifespan. We learned from the interviewees that some records contained very detailed information, and some were scarce. We also learned that depending on the criticality of the events, interest in the records varies on average between 24 hours and 30 days. If the interviewees needed to find any information, they applied available filters to the logged records and read through a filtered list. Occasionally, they use a string-match search to find records containing specific keywords they have previously seen in the text.

\begin{figure}[ht!]
\centering
\includegraphics[width=1.0\textwidth]{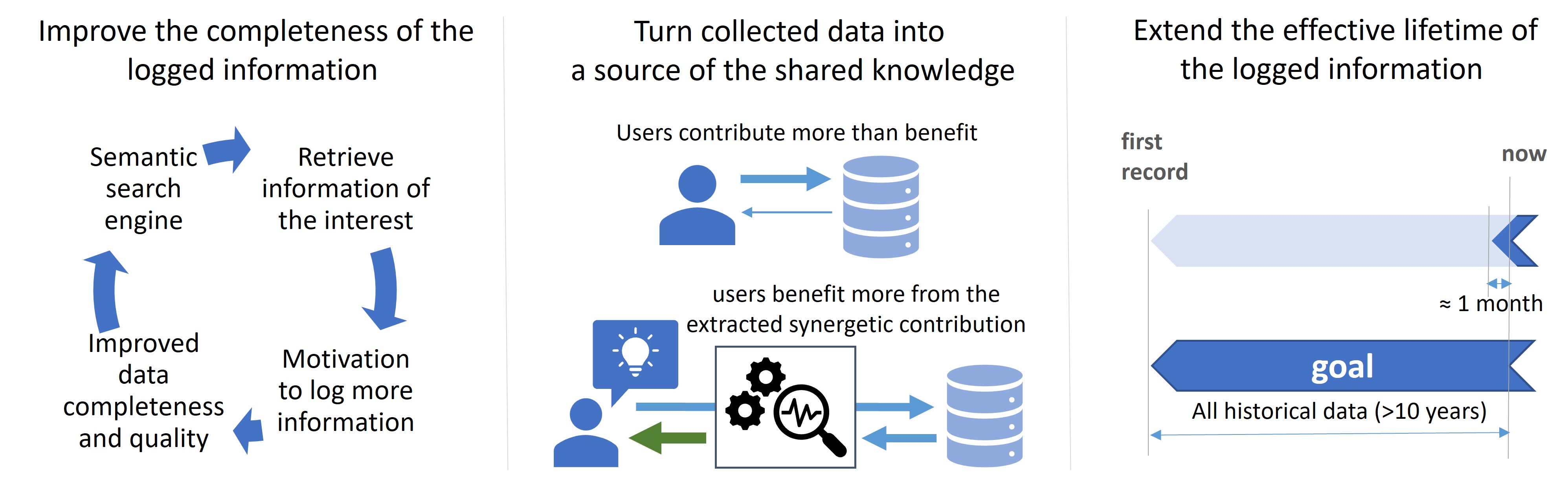}
\caption{The historical data of the plant operation contains a lot of know-how information on daily operations, but the use of it is quite limited due to the inefficient search. By implementing efficient semantic search and allowing users to find more complex records, the project aims to motivate the logging of more detailed descriptions of daily operations, thus improving the completeness and quality of input data.} 
\label{fig:motivation}
\end{figure}

\paragraph{Concept development with a click-dummy prototype}
Our goal for the NLP application was to improve the exchange of information, i.e., logging and retrieving crucial information. We expected the effective and efficient information exchange to yield higher production efficiency and shorter plant downtime. From the interviews, we learned about the common problems of data completeness, varying text quality, and the short life span of the records. To align with this goal and address the identified problems,  we proposed turning the logging system of daily operations into a source of shared domain knowledge (\Cref{fig:motivation}). As of the current state, only the most recent logged information is important for the users. We assumed that an NLP-driven domain knowledge processing system would enable leveraging the collected knowledge over the long run. Ultimately, users should be encouraged to log more information and improve their domain expertise by learning from the logged experience. 

As a first stage, we proposed a semantic search to enable efficient and effective access to the full history of daily operations. Our hypothesis was an efficient search engine for records should motivate users to log more complete and structured information. If the unstructured records were retrieved effectively, the new records with clear and complete information would have more confidence to be found later on demand. \Cref{fig:motivation} depicts a goal of improving the data completeness in the system by logging more information, e.g., from personal oral and written communication. If the discussions' key points can be later retrieved on demand, users get more motivated to transfer this information into the logging system. Representation of the domain language was a core challenge for this domain NLP application. For the prototype phase, we required a feasible solution to design, implement, and test within a short period but also efficient and robust to perform user studies for the prototype evaluation. \Cref{sec:prototype} reports more detail on this consideration.

To evaluate our concept, we performed Wizard of Oz prototyping \citep{browne2019wizard,dscoutUserResearch2019}, i.e., developed a ``click-dummy'' prototype, which mimics the functionality of the front-end of a semantic search engine. We adhere to the UX principle of ``fail fast'' and collect feedback on the proposed concept in the earliest possible stages of development. 

\begin{figure}[ht!]
\centering
\includegraphics[width=1.0\textwidth]{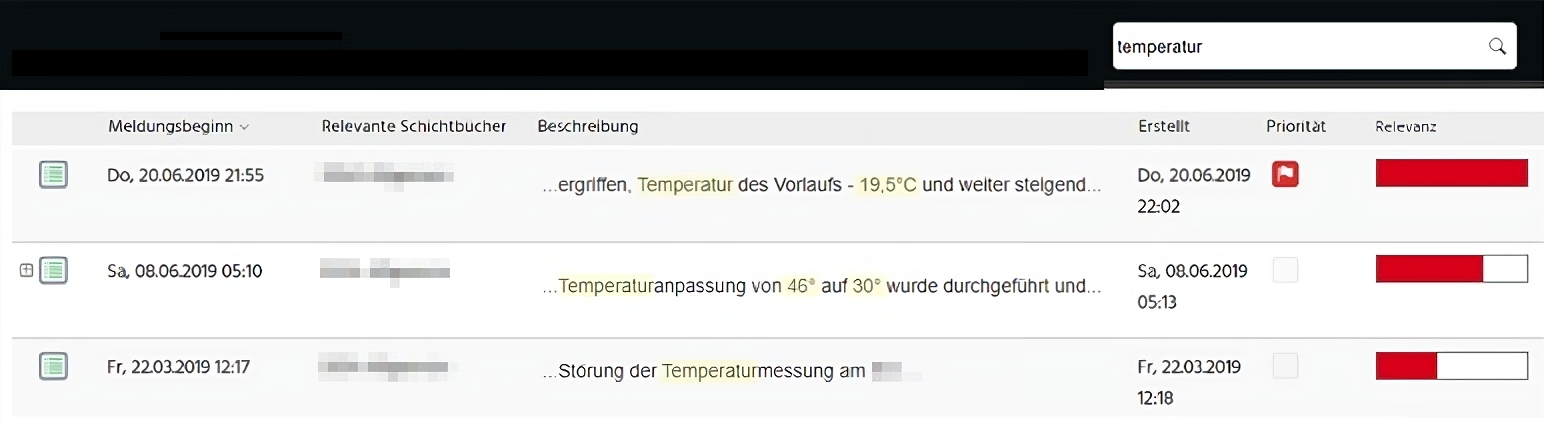}
\caption{Based on the opportunities learned from the user interviews, we created a ``click-dummy'' prototype for a moderated concept evaluation through user interviews (Wizard of Oz prototyping). We demonstrated this prototype to the interviewees and inquired if and how they would use it.} 
\label{fig:click_dummy}
\end{figure}

A click-dummy prototype aims to follow two users' mental models: the logging system and commonly used search engines, e.g., Google. \Cref{fig:click_dummy} shows that the prototype contains the visual attributes of a search field, a list of the resulting records, and such additional features as highlighting the matching excerpts and a similarity score. Moreover, its overall visual design is familiar to the users by resembling the overall design of the logging system. 

\paragraph{Moderated concept evaluation}

We conducted a second round of interviews where we presented the click-dummy prototype and evaluated our concept via guided interviews and moderated usability sessions. In the one-to-one interviews, we aimed to learn if providing efficient access to historical data would allow performing daily tasks in a more informed and effective manner:  
\begin{enumerate}
    \item Would an effective search engine assist you daily? If yes, how and when? Which queries would you devise? 
    \item Imagine that you have a system that could provide effective access to the expertise and shared knowledge about plant operations. Such a system could answer domain questions, extract information, and identify patterns in the text data. Which inquiries would you have for this system?
\end{enumerate}

All interviewees agreed that efficient search would enable them to access the previously logged knowledge, e.g., solutions to problems. Moreover, they highlighted that a search application would provide additional assistance on a night shift and could assist in planning product launches. The shared ideas for the queries helped us to create lists of queries for the later prototype evaluation (see \Cref{sec:eval}). \Cref{fig:interviews}, the last column, overviews the interviewees' replies on how they would use a system of shared domain knowledge. For example, they highlighted a need to access solutions to previously occurring problems. Moreover, they are expected to retrieve information with a deep understanding. For example, for a query ``quality state of product\_A on Monday,'' they would expect to see information about the batch level on that day. Additionally, they would explore information in an aggregating way, e.g., assemble problems that occurred to a specific product.

\subsection{Prototyping}
\label{sec:prototype}

The exploration phase resulted in a decision to develop a semantic search engine prototype for documents with uneven text structure and domain-specific language. The prototype's main challenge is handling the domain-specific language, including professional terms, abbreviations, shortenings, chemical formulas, various IDs, and plant machinery and equipment inventory codes. For the prototype development, we followed \textit{a UX principle ``fail fast, fail often''} \citep{dove2017ux} and implemented simpler approaches that require less time and resources. 

\subsubsection{Planning}
\label{sec:pipelines}
\paragraph{MVP requirements for the prototype development}
From the exploration phase and workshop, we deduct the following main requirements for the prototype: The search engine of the prototype shall be designed in a typical IR process architecture: (1) indexing of text records from a SQL database, (2) retrieving relevant records to a given query\footnote{We use the terms ``documents'' to refer to the records of the logging system, i.e., short documents.}. The most challenging requirement for the MVP is to tackle the problem described in the domain-specific language and technical style of the text records. It should use context expansion (i.e., adding definitions and descriptions of technical codes to the documents) and encoding documents vector word representation capable of handling out-of-vocabulary domain words, e.g., fastText. Furthermore, the MVP is supposed to be seamlessly integrated into the main software application.

\begin{figure}[ht!]
\centering
\includegraphics[width=0.8\textwidth]{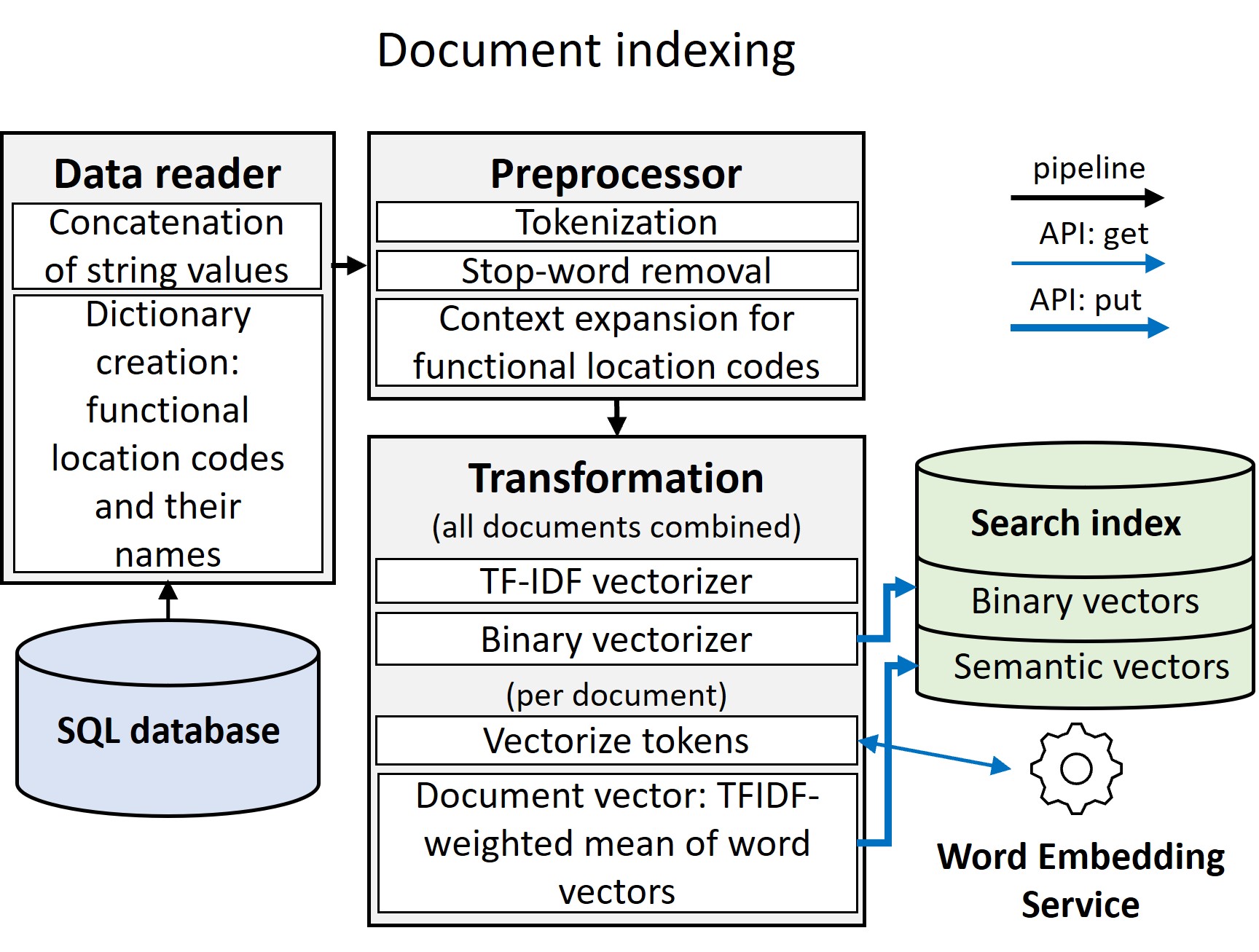}
\caption{Indexing text data from a SQL database is a process that includes data reading, preprocessing, data transformation through vectorization of words and documents, and saving the vectors into a search index. } 
\label{fig:indexing}
\end{figure}

\begin{figure}[ht!]
\centering
\includegraphics[width=0.8\textwidth]{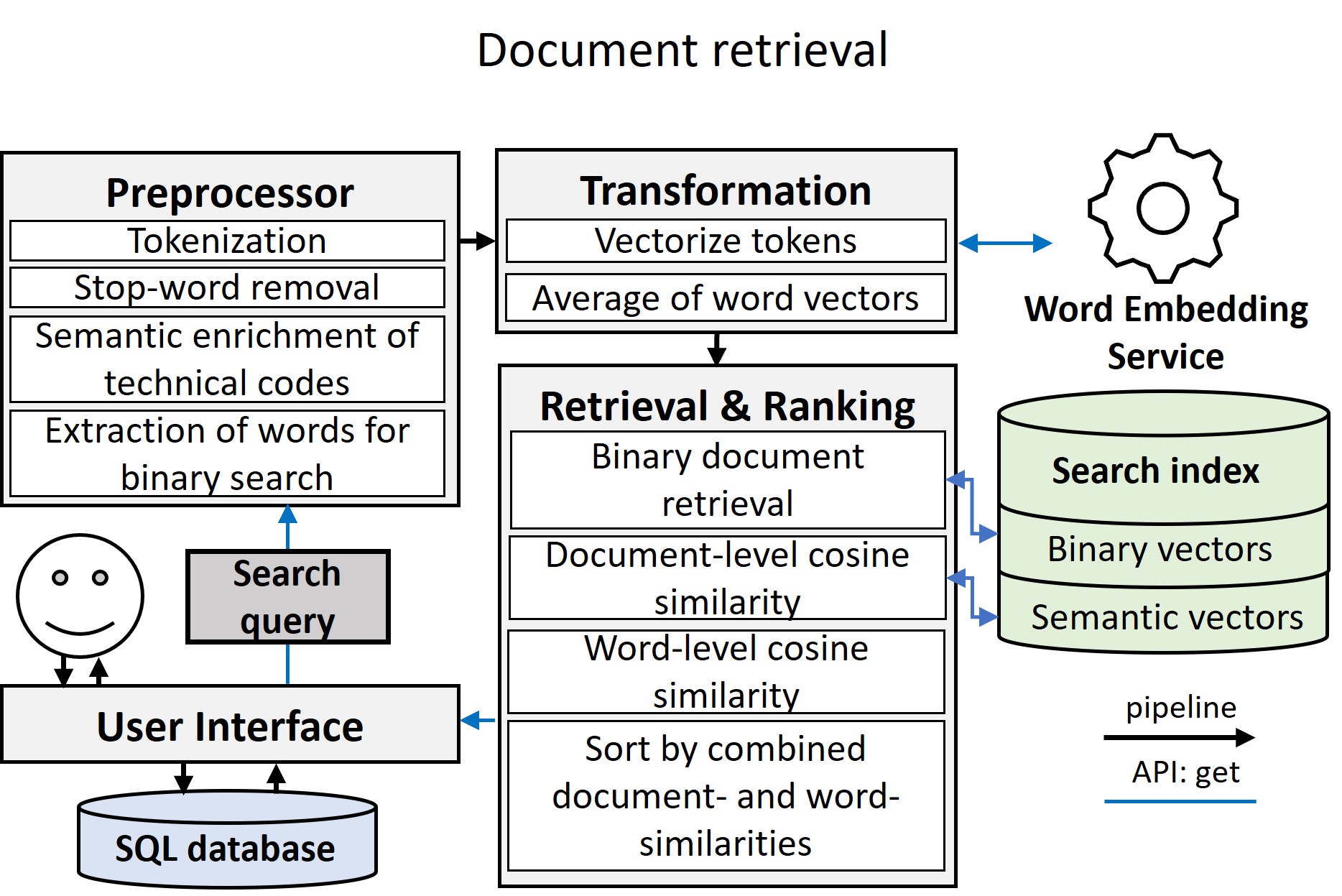}
\caption{Record retrieval preprocesses a search query, transforms it into a vector representation, retrieves documents based on binary retrieval and semantic similarity, and ranks the documents based on a combined document and term similarity.} 
\label{fig:retrieval}
\end{figure}

\subsubsection{Development - Backend \& Frontend}
\paragraph{Indexing}

Indexing starts with accessing an SQL database and reading values from string columns (\Cref{fig:indexing}). We concatenate all text fields of each record into one text document. Preprocessing performs \textit{context expansion} to add semantic descriptors to some domain terminology. We transform documents into multiple vector spaces: (1) vectorization into boolean and term frequency-inverse document frequency (TF-IDF) representations that require full-text collection, (2) per-document vectorization into semantic representation, which employs word embedding service to represent words into fastText word vectors and computes TF-IDF weighted-mean word vectors to represent documents. Finally, we save boolean and semantic document representations into a search index. Document retrieval uses the same word embedding service and represents a query as an average of word vectors. First, candidate retrieval retrieves documents that contain terms to be mandatory present in the documents (depending on if a query contains terms in quotes). Second, it filters the documents by semantic similarity (\Cref{sec:retrieval}). Lastly, we rank the top M most semantically similar documents by combined semantic similarity on the document and term levels.

\paragraph{Preprocessing}
\label{sec:preproc}
Preprocessing focuses on (1) providing additional context for some domain-specific terms, i.e., dictionary-based context expansion, and (2) normalizing text for semantic search that employs static word embeddings, e.g., lowercasing, stopword removal, and lemmatization. 

To enable context expansion, we create a dictionary about functional locations, i.e., production machinery and equipment installed in a plant. The dictionary contains information about long IDs, their short equivalents, and descriptions. We created this dictionary based on a graph structure of the machinery from the logging system.

We use the domain dictionary to extract descriptions of a mentioned machinery, e.g., in \Cref{fig:log}, a description of ``R105.12'' is ``reactor.'' The dictionary serves to ``translate'' machinery codes. 
Descriptions of the codes are added to the text before their occurrence. 
This approach is similar to automatic query expansion when query formulation is improved with additional information \citep{carpineto2012survey}. Such context expansion enables retrieving records by containing either specific codes of functional locations or their descriptions.

\paragraph{Representation of the domain-specific language}
\label{sec:transf}
Our semantic search is based on fastText word vectors, i.e., a skip-gram model for distributed representation on the char n-gram level \citep{mikolov-etal-2018-advances, bojanowski-etal-2017-enriching, grave2018learning}. Most of the word embedding models, such as word2vec \citep{mikolov2013distributed} and GloVe \cite{pennington-etal-2014-glove}, contain only vectors for the words that were in a training corpus. Out-of-vocabulary (OOV) words pose a big problem when word embeddings are applied to specific domains, e.g., biology or chemistry \citep{galea-etal-2018-sub}. Some models propose training word embeddings from scratch on a domain-specific corpus (cf. \citep{Efstathiou2018}) or using subword information to learn the representation of the words' parts to be able to represent later OOV words of this domain (cf. \citep{zhang2019biowordvec}). But we used a pre-trained fastText model because the vectorization on the n-gram level was sufficient to represent German compound terms of this domain\footnote{For example, the vocabulary of a trained fastText model does not contain the word ``Temperaturschwankungen'' (i.e.,  ``temperature fluctuations''). However, a fastText model can still represent this word so that it is semantically close to the words ``Temperatur'' and ``Schwankungen.'' Moreover, a shortening such as ``Temp.Schwank.'' will be vectorized and be semantically similar to ``Temperaturschwankungen.'' }. 
We use the TF-IDF-weighted average of the word vectors to highlight the impact of the semantically most important tokens. Weighting word embeddings with IDF has successfully been applied for IR tasks \citep{de2015learning, Galke2017, schmidt2019improving, jalilifard2021semantic, arroyo2019unsupervised, agarwal2019authorship}.  

We did not use contextualized word embeddings, e.g., language models, because word representation may suffer from incomplete grammar or syntax of the domain short texts in our dataset. Therefore, we used simpler static word embeddings, which can more robustly represent the writing style of hand-written notes and do not require word sense disambiguation within the same narrow domain. Later prototype improvements will be based on the domain adaptation of the dense passage retrieval based on sentence transformers \citep{reimers-gurevych-2019-sentence, wang-etal-2022-gpl, guo2022domain, thakur-2021-BEIR, wang-etal-2021-tsdae-using}.

\paragraph{Retrieval and ranking}
\label{sec:retrieval}
Document retrieval consists of two parts: semantic search for regular words and exact match for any digit-containing terms (or on demand). While semantic search accommodates synonyms and modification of word forms, exact match enforces specific criteria. Such an approach defines how specific search results should be. For example, a user can search for temperature deviation in general or an inventory code of a functional location.

From the candidate documents, we retrieve documents that are the most semantically related to a query. Candidate retrieval computes document similarity as cosine similarity between a vector representation of a query and vector representations of all candidate documents. A linear search is not time-efficient, but in the current prototype phase, higher accuracy of the approach is more important than execution time. Search in the domain of daily operation is similar to Web search, i.e., only a few of the most relevant records are important for the users. Therefore, we retain the top $K=200$ most similar documents and proceed with ranking. 

\begin{figure}[ht!]
\centering
\includegraphics[width=1.0\textwidth]{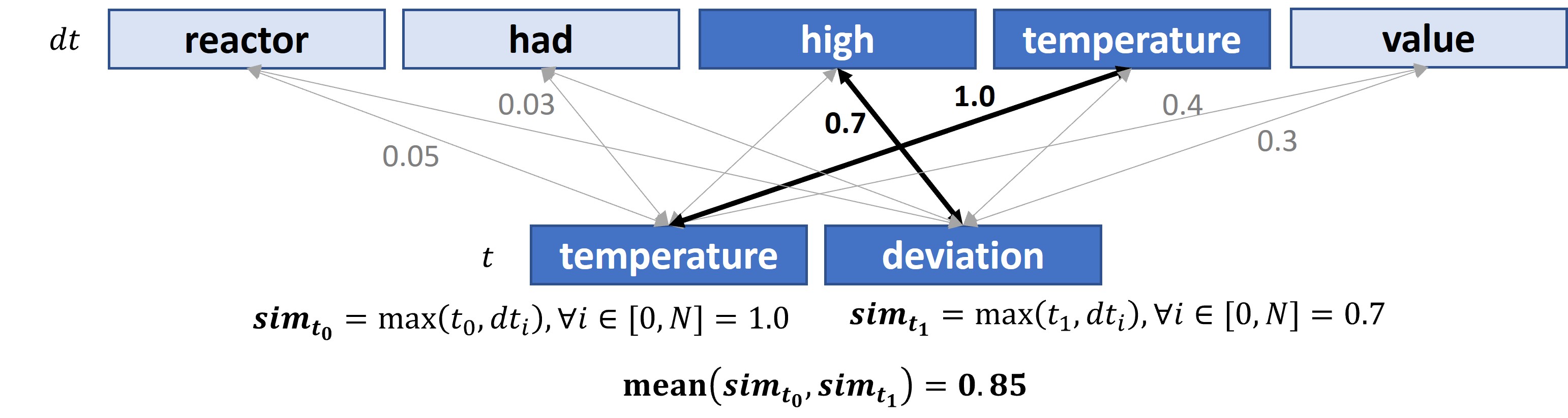}
\caption{Term-level similarity identifies the most semantically similar terms between a document and query.} 
\label{fig:term_sim}
\end{figure}

Our ranking is based on a harmonic mean of document and term similarities. Term similarity prioritizes documents containing the query's terms as-is and ranks the documents containing synonyms higher than the query's terms. Term similarity is computed as a mean of the maximal similarities of the query's terms to the document's terms (\Cref{fig:term_sim}): 

\begin{flalign}
TS_j = \frac{\displaystyle\sum_{qt \in QT} \operatorname*{max}_{ dt \in D_j} cossim(qt, dt)}{|QT|}
\end{flalign}
where $QT$ is query terms,  $|QT|$ is a number of query terms, $qt$ is a vector representation of a query term, $D$ is a document, and $dt$ is a document term. A harmonic mean balances general semantic relevance on the document level and precise similarity on a term level that allows synonyms or shortenings of the terms to gain a high rank. If an exception situation occurs that a query contains only terms for an exact match, then the documents are ranked by their timestamp.

\begin{figure}[ht!]
\centering
\includegraphics[width=1.0\textwidth]{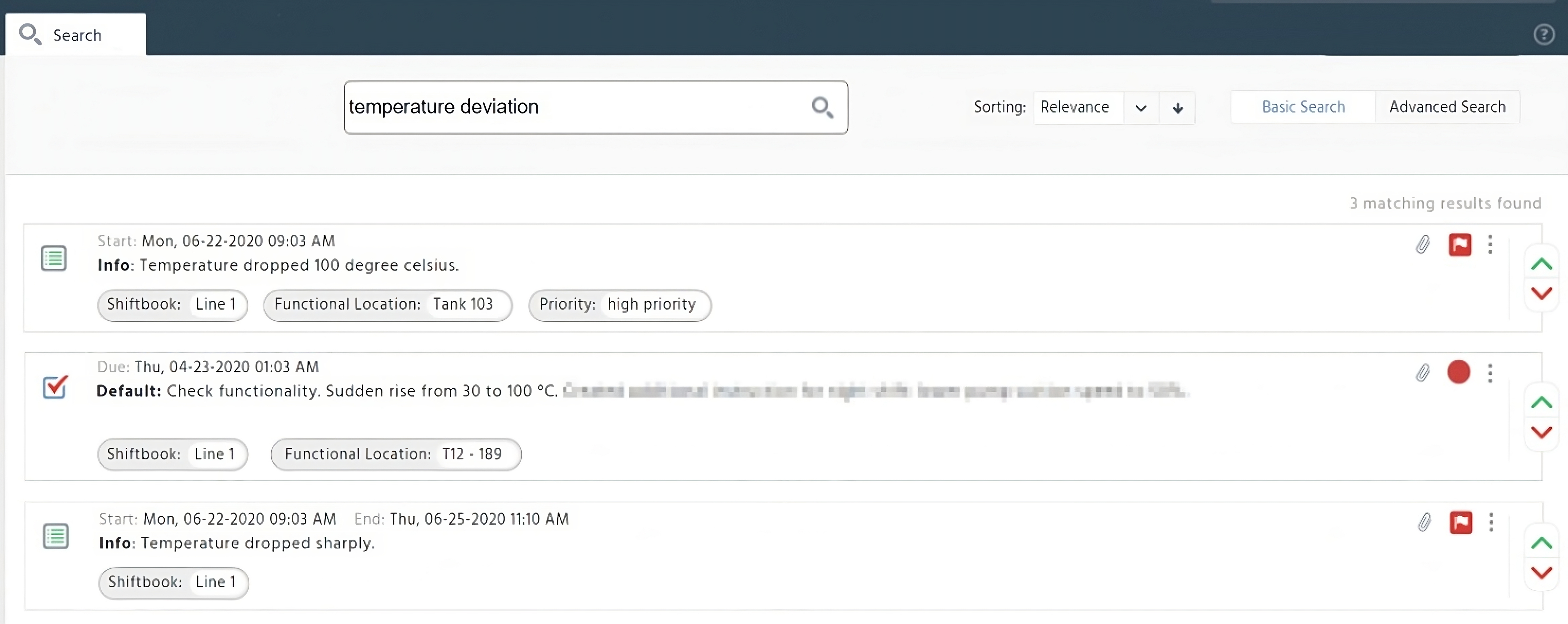}
\caption{User interface of the prototype resembles the proposed concept (\Cref{fig:click_dummy}). The prototype contains functionality for the relevance feedback for each record but does not contain previously proposed visual elements that indicate computed relevance. This is made to prevent bias in the user assessment during user studies. } 
\label{fig:alpha_prototype}
\end{figure}

\paragraph{User interface}
The user interface of our prototype resembles a click-dummy prototype proposed for the concept evaluation. \Cref{fig:alpha_prototype} depicts an interface of the prototype that contains the text of the record, enables sorting by relevance and timestamp, and contains functionality to collect user feedback for the prototype evaluation. Additionally, the prototype provides a detailed view by clicking on a record of interest. For the first version of the prototype, we excluded the functionality of highlighting and visualization of the relevance score to prevent judgment bias during the user-based evaluation of the prototype accuracy (see \Cref{sec:acc}).

\subsection{Evaluation}
\label{sec:eval}
Our user-driven prototype evaluation aims to assess how users utilize the prototype in their work routine, i.e., how the effectiveness of their daily tasks changes after the unmoderated interaction with the prototype, and how accurately the prototype performs. We recruited a focus group of seventeen domain experts and advanced users of the logging system from one plant for the prototype evaluation with a diverse set of roles to ensure opinion variance in the user studies (\Cref{tab:alphatest_participants}). Some study participants were previously recruited for the exploratory generative UX research (see \Cref{sec:exploration_interviews}). The users were asked to perform two types of interaction: (1) freely use the system as they see fit during their work tasks (\textit{user-driven prototype evaluation}), (2) perform a set of tasks to assess the relevance of search results (\textit{accuracy evaluation}).

\begin{table}[ht!]
\caption{Demographics of the user study participants for the prototype evaluation. All participants are domain experts and advanced users of the logging system.}
\label{tab:alphatest_participants}
    \centering
    \begin{tabular}{|c|c|c|c|}
    \hline
    Role & \makecell[c]{Invited to the \\prototype evaluation} & \makecell[c]{Performed relevance \\assessment \& \\Questionnaires} & \makecell[c]{Participated \\in interviews}  \\ 
    \hline
    \makecell[c]{Shift leader} & 10 & 7 & 4\\
     \hline
    \makecell[c]{Second plant manager} & 2 & 2 & 1\\
     \hline
      \makecell[c]{Production assistant} & 2 & 2 & 1\\
     \hline
     \makecell[c]{Production \& Maintenance} & 2 & 2 & 1\\
     \hline
     \makecell[c]{Laboratory manager} & 1 & 1 & 1\\
     \hline
     \hline
    \makecell[c]{\textbf{Total}} & 17 & 14 & 8\\
     \hline
    \end{tabular}
\end{table}

The section is structured as follows. \Cref{sec:acc} reports the relevance assessment and evaluation according to the information retrieval (IR) methodology. \Cref {sec:experience} summarizes questionnaires and user interviews that measure differences in the user experience before and after interacting with the prototype.  

\subsubsection{User-driven prototype evaluation}
\label{sec:experience}
UX evaluation estimated how user utility and effectiveness change before and after using the prototype, e.g. if users found the information more efficiently and effectively. The users were asked to complete a questionnaire and were interviewed before the unmoderated interaction with the prototype (i.e., control point). After several weeks of free use of the prototype, we repeated the questionnaires and interviews to reflect on how their pattern of acquiring information has changed.  Additionally, through interviews, we collected free-form feedback on using the prototype and ideas for improvements from a user perspective. Below, we report quantitative analysis on the questionnaires and qualitative analysis of the interviews.

\paragraph{Questionnaires: Control point and reflection} 
The questionnaires aimed to compare two ``snapshots'' of the user experience before and after the prototype use. We wanted to see how the study participants obtained information by using the prototype compared to old methods of the logging system when only the default search had been available. 

\begin{figure}[ht!]
\centering
\includegraphics[width=0.85\textwidth]{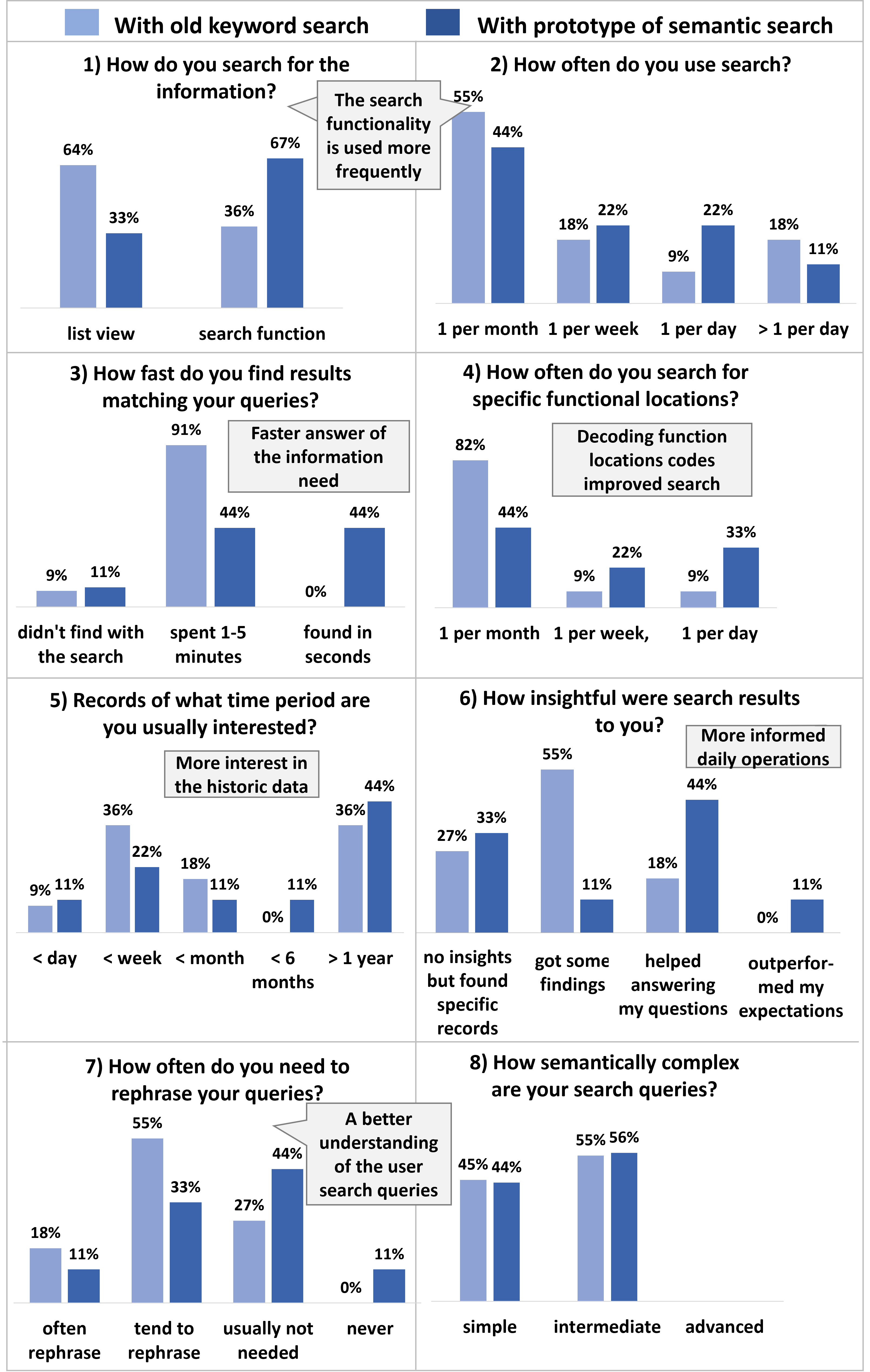}
\caption{Comparison of the user experience in searching for information in the system. The prototype showed a positive effectiveness and user utility trend using historical data to assist their daily work.} 
\label{fig:user_exper}
\end{figure}

\begin{table}
\centering
\caption{Summary of the questions from the user interviews before and after using the prototype about the tool preference to search for the information. Although the interviewees continued using list view for specific tasks, semantic search became a preferred method of finding information of interest.}
\label{tab:user_studies}
\begin{tabular}{p{0.08\textwidth}|p{0.35\textwidth}|p{0.45\textwidth}}
\hline
 & \makecell[c]{\textbf{List view}} & \makecell[c]{\textbf{Search functionality}} \\
\hline
\makecell[c]{\rotatebox[origin=c]{90}{\textbf{Before prototype}}} & \makecell[l]{- The search is not efficient \\enough \\
- Got used to the list view (x2)\\
- Need to perform comparison of \\multiple entries and/or specific \\fields (x2)\\
- Don’t know any other way \\
- The fastest solution 
} & \makecell[l]{(\textbf{Keyword search})\\ 
- Use a search to filter out a lot of non-\\relevant entries \\
- A habit of googling \\
- Know terms for searching \\
- Determine redundancy in the records \\
- Find solutions
} \\
\hline
\makecell[c]{\rotatebox[origin=c]{90}{\textbf{After prototype}}} & \makecell[l]{- Search for specific information, \\ e.g., data of old analysis  \\ 
- Faster for some tasks
} & \makecell[l]{(\textbf{Semantic search})\\
- Faster and easier, especially when \\looking for the older  entries \\
- Easy to use \\
- Look for specific solutions, problems, or\\ completed tasks \\
- Find the results faster (x2) \\
} \\
\hline
\end{tabular}
\end{table}

\Cref{fig:user_exper} reports the aggregated results of the questionnaires. For each question in \Cref{fig:user_exper}, the answers are sorted from the worst to the best option. After using the prototype, we observed a shift in the answers towards higher values for all eight questions. We summarized the improvements made by the prototype as follows. (Q1) Before the participants accessed the prototype, they read through a list of logs over the past 24-72 hours instead of using the simple search functionality. (Q2-Q3) We observed that the frequency of search functionality also increased, probably due to the speed-up and better search results of finding answers to their questions. (Q4) The functionality of the context expansion facilitated more efficient search and, consequently, frequent searching for information about them. (Q5) The ultimate goal of the project was to increase the effective lifespan of the logged information (see \Cref{fig:motivation}). The chart depicts that after the introduction of the semantic search, the study participants showed more interest in the historical data over one month. (Q6) Performing daily operations became more informed due to the increased number of cases when search results helped answer the questions. (Q7) The study participants indicated they needed to rephrase their queries more rarely than before to obtain expected results. (Q8) Although we saw a marginal increase in the complexity of the search queries, the quality of the search results improved (see other questions). We expect the numbers to be more pronounced if operators use semantic search in a production system and create more semantically complex queries.

\paragraph{User interviews: Control point and reflection} 
User interviews with open-ended questions helped in reflecting on the developed prototype. We collected feedback from interacting with the prototype and ideas for improvements in the next version: 
\begin{enumerate}
    \item How was the experience with the semantic search? How helpful was it? Did it increase your performance?
    \item How did you feel about using the prototype? Excited? Curious? Disappointed?
    \item Did you search with your search queries? Were there any search queries for which you would have liked to have found more? Was there a search query for which you were particularly impressed by the results or for which you found interesting information?
    \item Imagine the prototype is deployed into the productive system. Which functionality would make a difference in using the semantic search regularly? What could be improved? 
\end{enumerate}

\Cref{tab:user_studies} reports summaries of the eight interviews and compares finding information by reading records in a list view to using search functionality, i.e., keyword vs. semantic search. Before the prototype evaluation, most interviewees preferred using a list view as it was a fast method that complemented the inefficient old keyword search. Despite low functionality, the old search was still used to find records by keywords. According to some interviewees, using any search engine felt natural due to regular interaction with the Google search engine. After the pilot, on the one hand, a list view was still preferable for specific tasks, i.e., data comparison. On the other hand, semantic search enabled more effective and efficient access to historical data. According to some interviewees, semantic search was a helpful tool to find solutions to similar problems accounted for in the past. 

\subsubsection{Accuracy evaluation}
\label{sec:acc}
The accuracy evaluation focused on the precision-oriented IR evaluation and estimated the relevance of the $N$ top-ranked results. Although such ``shallow'' evaluation could be incomplete, it is consistent if users are interested only in top-ranked documents in the domain application \citep{LiuOezsu2009}. Therefore, the relevance assessment enabled the estimation of the prototype's accuracy for daily use.

\paragraph{Collecting a test dataset} 
For the user evaluation, we prepared a list of 100 query topics. We obtained ideas for the query topics during the concept evaluation phase (see \Cref{sec:exploration_interviews}) and created the queries based on the plant's logging system records. By following the methodology of \cite{clough2013evaluating}, we collected diverse queries, phrases of which occurred in the system database at least once. The queries differ by length of query topics, i.e., a query contains 1-4 search terms. Moreover, the queries contain diverse domain terms such as machinery IDs, processes, statuses, and repair steps. 

\begin{figure}
\centering
\includegraphics[width=0.7\textwidth]{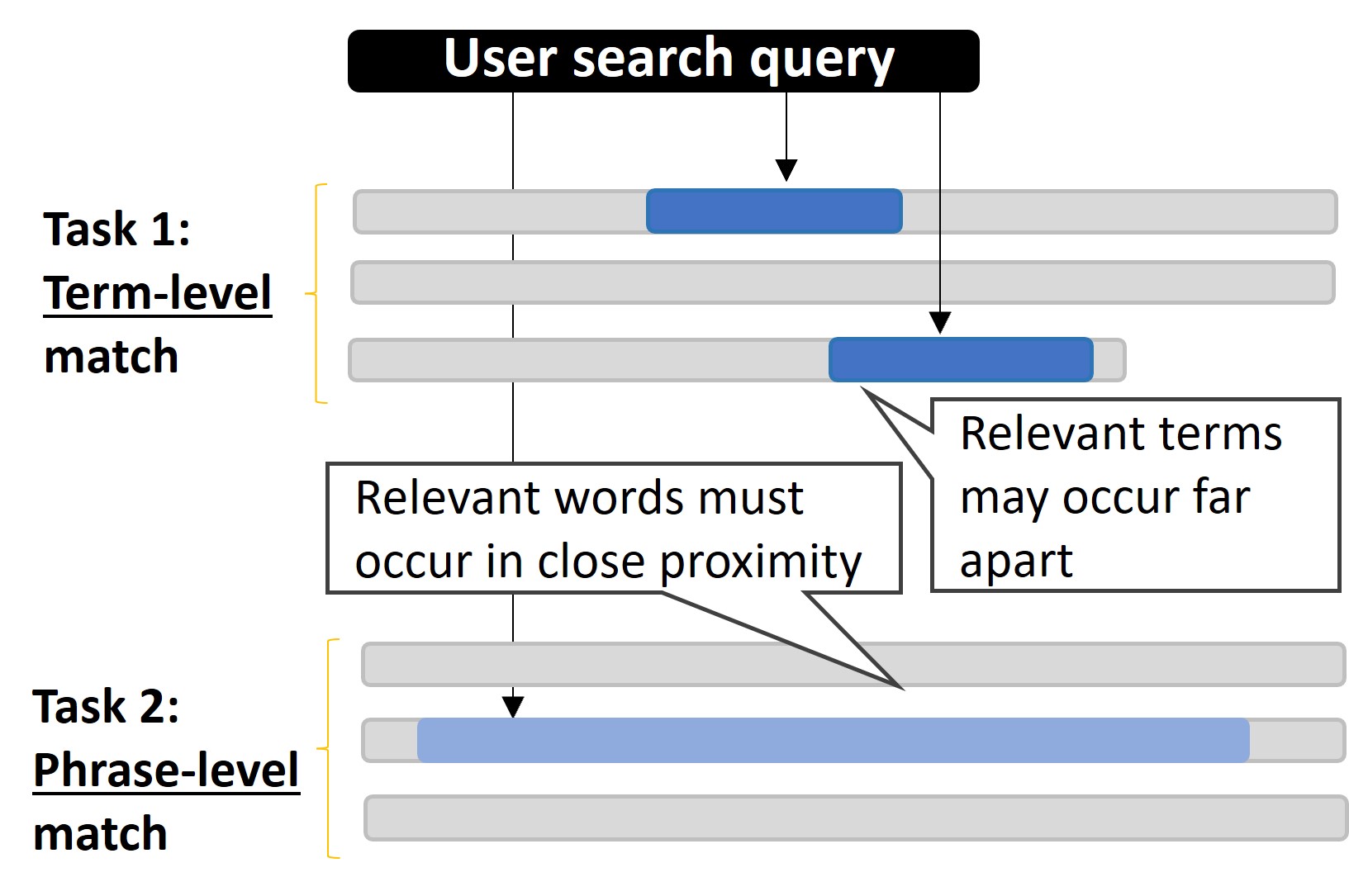}
\caption{Study participants needed to perform a set of tasks in the prototype. The tasks included assessing document relevance on two levels: term and phrase match. The difference between the tasks is in the proximity of the relevant terms in a document.} 
\label{fig:task}
\end{figure}

\paragraph{Controlled task execution} 
Each of the fourteen participants received 20 queries for evaluation and needed to assess the relevance of the search results on two levels: term and phrase (\Cref{fig:task}). We distributed a list of query topics among the participants so that at least two participants evaluated one query topic. 
The same queries needed to be executed twice to assess the relevance of each of the 20 top-ranked results on the two levels. For a term level, a document was to be labeled as relevant if all query terms occurred in a document as-is or as a synonym independently from their word order or location within the text. For a phrase level, a document was relevant if a query phrase or its meaning occurred in close proximity, i.e., in a narrow context. For example, for a query topic ``problem with a pump,'' the documents that described a problem without explicitly mentioning the word ``problem,'' e.g., ``a leakage in a pump,'' were to be annotated as relevant.

We calculated inter-rater agreement between the assessors who had evaluated document relevance for the same queries to ensure the experiment's validity. The assessors evaluated approximately half of the distributed queries, i.e., 44 queries for the first task and 16 queries for the second task were assessed at least twice. We calculated Cohen's kappa for each task \citep{mchugh2012interrater} to measure inter-rater agreement. On the term level, \Cref{tab:kappa} reports $\kappa_{term} = 0.856$, a near-perfect agreement between the study participants. On the phrase level, $\kappa_{phrase} = 0.673$, i.e., the study participants reached a substantial agreement. We assume that lower inter-rater agreement on the phrase-level document relevance depends on two factors: (1) an insufficient number of assessed queries for the second task yielded less reliable kappa computation, (2) assessing phrase-level relevance remains a harder task due to ambiguity in the task description, i.e., what is close proximity. For example, a relevant term of a functional location's code was at the beginning of a record, whereas the remaining relevant terms occurred later in the document. Therefore, a setup could be unclear on how to assess.

\begin{table}[ht!]
\caption{Inter-rater agreement of Cohen's kappa for two tasks: term and phrase level similarity. The level of agreement for both tasks is substantial.}
\label{tab:kappa}
    \centering
    \begin{tabular}{|l|l|l|l|l|l|l|}
    \hline
        \makecell[c]{Document relevance level} & \# queries & $\kappa$  \\ 
        \hline
        term level  & 44 & .856  \\ 
        phrase level  & 16 & .673   \\ 
        \hline
    \end{tabular}
\end{table}

To create a test dataset, we combined the assessors' relevance votes for the query topics assessed more than two times. For each query, we took a list of search results. We summed up votes across assessors and tasks and assigned the results as document relevance scores. For example, if two assessors indicated that a document was relevant on both term and phrase levels, then a document got a relevance score of 4. 

\paragraph{Computing accuracy metrics} 
We evaluated the prototype and the baselines against the documents collected during the expert assessment of the prototype. We acknowledge that such evaluation may cause certain biases favoring the prototype over baselines. 
But for the precision evaluation of the prototype, we consider such accuracy approximation sufficient to show how user utility improved by using the prototype. In our evaluation setup of a few user study participants, A/B testing would have been inappropriate and could have caused statistically irrelevant results \citep{gilotte2018offline}. 

After implementing the prototype, we provided two baselines to evaluate how user utility changed: a keyword search (i.e., exact term match) and BM25. The keyword search is a default SQL-based search from the logging system with which the domain users are familiar. BM25 represents a standard baseline used in IR. 
We implemented a more flexible version of the keyword search for the experiments. 
A document was retrieved if at least one term from a query was found. The documents were ranked by the decreasing number of overlapping terms and then by a timestamp of the records. 
For BM25, we normalized German text records and accommodated local domain setup with lemmatizing and removing stopwords. 
We retrieved documents if cosine similarity was higher than 0.15 and ranked them using it and a document's timestamp.
We implemented an optional context expansion in both baselines to evaluate its impact on handling the domain-specific language (see \Cref{sec:preproc}).

We evaluated the accuracy of the prototype at top $N$ records with precision-oriented metrics, i.e., precision $P$@N, and metrics for ranking quality, i.e., mean average precision $mAP$@N \citep{LiuOezsu2009}, mean reciprocal rank $MRR$ \citep{radev-etal-2002-evaluating} and normalized discounted cumulative gain $nDCG$@N \citep{jaervelin2002cumulated}. The metrics were calculated for the top 5 and 20 records in our experiments. The combined relevance scores in the test dataset were used to compute $nDCG$@N.

\begin{table}[ht!]
\caption{Results of the relevance assessment. The proposed prototype of semantic search significantly outperforms all baselines.}
\label{tab:eval_results}
\footnotesize
    \centering
    \begin{tabular}{|l|c|c|c|c|c|c|c|c|c|c|}
    \hline
        \makecell[c]{Search \\method} & \makecell[c]{Context \\expans.} & \makecell[c]{Avg. \# \\retr. \\docs} & $MRR$ & $P$@5 & \makecell[c]{$mAP$\\@5}  & \makecell[c]{$nDCG$\\@5} & $P$@20 & \makecell[c]{$mAP$\\@20}  & \makecell[c]{$nDCG$\\@20}\\ 
        \hline
        \multirow{2}{*}{\makecell[l]{keyword \\search}}  & - & 3300 & .153 & .068 & .125 & .076 & .063  & .112  & .096 \\ 
                                       & + & 5944 & .132  & .180 & .077 & .152 & .101  & .066  & .100\\ 
        \hline
        \multirow{2}{*}{BM25}        & - & 736 & .582 & .294 & .548 & .322 & .178  & .479 & .291 \\ 
                                       & + & 657 & .572 & .302 & .523 & .337 & .182  & .459 & .291 \\ 
        \hline
        \makecell[l]{prototype: \\ semantic \\search}   & + & 90 & \textbf{.888} &\textbf{.749} & \textbf{.880} & \textbf{.725}   & \textbf{.668}  & \textbf{.805}  & \textbf{.772} \\ 
        \hline
    \end{tabular}
\end{table}

\Cref{tab:eval_results} reports the proposed prototype's evaluation results and baselines. Our prototype outperformed all baselines, but even BM25 showed significant performance improvement compared to the original search method in the logging system. Compared to the keyword search, BM25 showed 3-4 times higher values in all metrics, whereas the semantic search improved evaluation by 5-10 times.

\begin{figure}[ht!]
\centering
\includegraphics[width=1.0\textwidth]{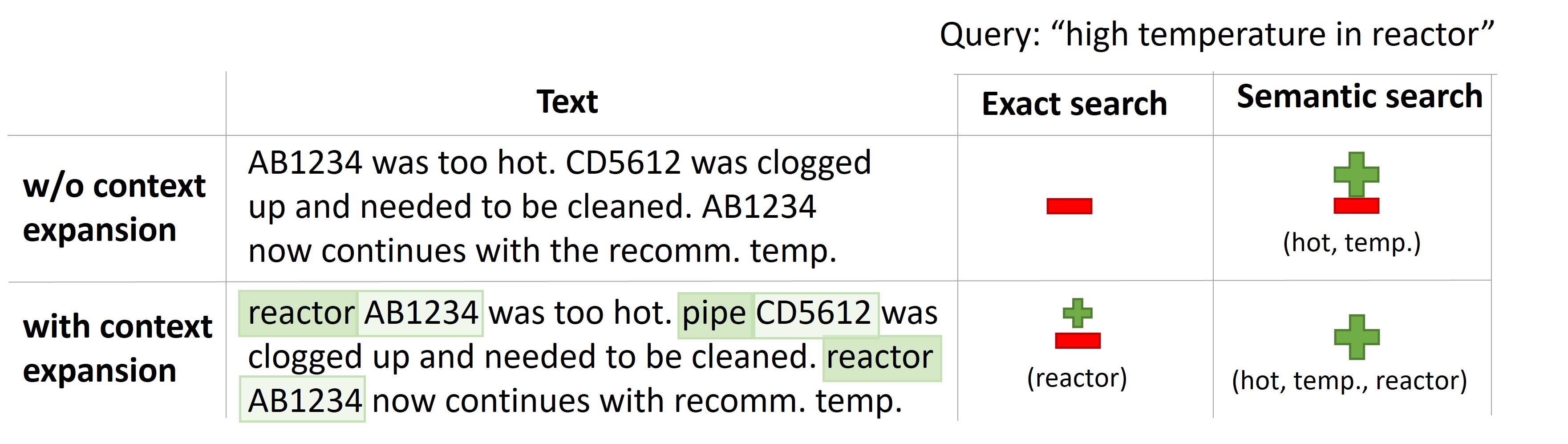}
\caption{An example of the positive influence of context expansion on all search methods in the domain language of daily operations in the process industry.} 
\label{fig:context}
\end{figure}

Using context expansion improved precision $P$@5 and $P$@20 of both baselines. \Cref{fig:context} depicts the influence of context expansion. The ranking values measured by $MRR$, $mAP$@5, and $mAP$@20 that ignore the annotated relevance score of the documents have slightly dropped. On the contrary, $nDCG$@5 and $nDCG$@20 have increased. The results show that context expansion assisted in retrieving more candidate documents and ranking relevant documents higher.

\section{Discussion}
\label{sec:discussion}
The paper reports a case study for which we devised and implemented a methodology of making generative UX research a part of a pipeline for prototyping NLP applications. The case study showed that learning directly from users or target audiences enabled the identification of their problems and needs, which consequently led to effective and efficient prototyping with positive user feedback in the evaluation phase. Therefore, instead of evaluating hypothetical ideas for an NLP application, the concept generation phase addressed these specific needs in a problem-solving way. 
Our case study took approximately 30 weeks to perform a full cycle to develop and collect extensive feedback on the MVP (4 weeks for exploration, 16 weeks for prototyping, and 10 weeks for evaluation). This period is rather short in terms of full-scale project development but yields better products tailored to the needs of domain experts.

We expect the proposed methodology of combining UX and NLP methodologies to positively impact domain-specific NLP application development. Understanding and evaluating data quality and availability must come in a tight combination with understanding the domain users and their stories. A timeline depicted in \Cref{fig:timeline} can serve as a step-by-step guideline for the generative-driven prototyping of NLP domain applications. The pipeline describes specific tasks for UX researchers, data scientists, and project managers at each stage. Combining the steps in each type of UX research and the questions from the user studies (\Cref{sec:exploration} and \Cref{sec:eval}) expands the guideline and equips the UX practitioners with applied examples. \Cref{sec:data_analysis}, \Cref{sec:exploration_interviews}, and \Cref{sec:prototype} exemplify data scientists how we performed exploratory data analysis, assessed and solved the challenges for a domain NLP prototype application.

We encourage NLP and UX researchers and practitioners to join in evaluating this methodology and reporting their results. Currently, we are testing the methodology in a second case study in which we prototype an NLP application for supporting domain users in logging information. There, we observe and explore how the domain users log information, which steps are hard for them, and which information they enter wrongly or incompletely. Although it is too early to present the results of the case study in full detail, they support our conclusion that using generative UX research for NLP applications strongly improves user acceptance.

\paragraph{Impact of generative UX research on the domain NLP applications}
Generative UX research focuses on contextual inquiries to learn about the daily routines of domain users and how they use an existing product (if any). Observing and exploring which challenges, concerns, and needs they have enables opportunities for a new NLP application/feature. To facilitate immersed domain understanding, contextual inquiries include interviews with the domain users and observations in their environment. Generative research engages domain users from the first step of the development process and targets the identified pain points. This problem-solving methodology increases interest in the final product and trust in the final application. In our case study, we applied generative UX research to learn how the interviewees use the logging system in their daily operations in the process industry domain. The interviews highlighted two problems: (1) the logged text information has an uneven reporting style and suffers from partial incompleteness, and (2) users mostly used only recently logged information and used inefficient ways to access historical data. The identified problems establish a ground for concept generation.

Exploratory data analysis is essential for ML/NLP applications to understand data availability and quality. In generative research, the preliminary data analysis enables learning about the data constraints that must be considered during the concept generation phase. Moreover, the data analysis facilitates forming hypotheses about patterns of special usage of a domain system, which can be later tested during user interviews. Our exploratory data analysis revealed multiple challenges of the domain text data: (1) semi-structured data records included structured attributes and unstructured text fields, (2) writing style in the records varied from large structured reports to notes quickly taken down with incomplete grammar and syntax, (3) domain language contained rare and domain special terminology, e.g., abbreviations, term shortenings, chemical formulas, numerical values, and codes of the machinery that act as domain terms.

Generated ideas and concepts from UX research help define the scale of a project. Selecting small ideas for a prototype phase speeds up evaluating the credibility of a bigger concept. Based on our UX research, we created an idea for a big-scale system for domain knowledge sharing, e.g., a solution recommendation system that is based on the experience of plant operation, collected and extracted from the historical records (see \Cref{fig:motivation}). We selected a small part from this idea for prototyping and proposed a semantic search to provide effective access to the historical data. The prototype provided us with quick feedback from the domain users and suggested steps for further development.

\paragraph{Prototyping for UX+NLP research: quick and effective}
Prototyping for NLP applications should aim at the UX principle ``fail fast, fail often.'' It means that a primary goal is to evaluate prototype credibility, which leads to a prototype containing only basic or simplified functionality sufficient for estimating the user utility. Moreover, industrial or domain NLP applications may not contain top-tier research because real-world applications need to deal with more constraints than academic research. Simpler methods do not always mean less effective but more robust and suitable to a specific domain or low-resource language. In our case study, we used a combination of two solutions to handle the domain-specific language in the prototype: (1) fastText word embeddings that reliably represent domain compound German terms and (2) context expansion that adds descriptions to the mentioned machinery codes. Although these solutions are not from the cutting-edge research of language models and their domain adaptation, the prototype enabled reliable evaluation of the user utility and usefulness of the proposed NLP application. 

\paragraph{Prototype evaluation: effectiveness of the daily task completion }
Prototype usefulness and user satisfaction in domain-specific NLP applications are often bound to the personal effectiveness of completing the tasks for which the prototype was intended. Depending on the application, effectiveness can be replaced or extended with efficiency. Since the effectiveness of completing tasks and user utility often depend on the accuracy of an NLP application, the evaluation needs to address a combination of prototype accuracy evaluation when the system is in use and user satisfaction in assisting in their tasks. Quantitative and qualitative analysis of accuracy metrics, questionnaires, and interviews can yield the most complete arguments to decide about the prototype's success or failure and points for its improvement.  

In our case study, user utility in the process industry depends on the effective and efficient acting on occurring problems, which yields safety and quality of the technological processes. 
The evaluation methodology employs domain assessors to estimate the accuracy of the prototype and its impact on their daily routine. First, we perform a precision-oriented evaluation of the prototype and ask domain users from a focus group to assess the top search results of a given list of queries. 
Second, we analyze user utility with questionnaires and personal interviews. Analysis of the questionnaire difference before and after the prototype evaluation showed a positive trend in improving user experience working with historical logged data.

\section{Conclusion}
The paper reports on a case study of developing a domain-specific NLP application prototype in the process industry. To develop a prototype, we proposed a methodology for integrating generative UX research into the data-driven methodology of developing NLP applications by reinforcing domain understanding through data analysis and user interviews.
Generative UX research focuses on domain users and employs them at the initial stages of prototype development, i.e., ideation and concept evaluation and the last stage of the prototype evaluation. The core idea of generative research is that involving domain users increases user utility, interest, and trust in the final application since it targets their needs. 

The case study reports a full prototype development cycle of a domain-specific NLP application for daily operations in the process industry. In the generative UX research, we learn about a complicated domain of the German language, varying writing styles, partial data incompleteness in the logging system, and the short lifetime of the text records. We propose a semantic search to enable effective access to the historical knowledge of plant operations and motivate users to improve data completeness by logging more information. The user-oriented prototype evaluation focuses on assessing the accuracy and user utility when interacting with the prototype with a set of provided tasks or openly as they see fit. The prototype evaluation demonstrates more informed, effective, and efficient task completion in daily operations.

The paper shows that generative research for NLP application development combines the strengths of both UX and NLP research. The methodology is a promising direction for broad use in the development of NLP applications. The UX+NLP combination reveals opportunities and constraints of the data- and user-driven nature through exploratory data analysis and contextual inquiries, which enables a more informed and effective development process. We encourage UX+NLP teams to work in close collaboration to test the proposed methodology and report their results. 

\section*{Acknowledgements}
This project is supported by the ZIM program (Zentrales Innovationsprogramm Mittelstand) run by the German Ministry of Economic Affairs and Climate Action (BMWK) on the basis of a decision by the German Bundestag.

\bibliography{sn-bibliography}%

\end{document}